\documentclass[10pt,journal,compsoc]{IEEEtran}

\usepackage{fontawesome}
\usepackage{hyperref}
\usepackage{tabularx}
\usepackage{graphicx}
\usepackage{amsmath}
\usepackage{amssymb}
\usepackage{url}
\usepackage{threeparttable}
\usepackage{framed}
\usepackage{multirow}
\usepackage{gensymb}
\usepackage{hhline}
\usepackage{rotating}
\usepackage{babel}
\usepackage{mathtools}
\usepackage{booktabs}
\usepackage{blindtext}
\usepackage[normalem]{ulem}
\usepackage[table]{xcolor}
\usepackage{lettrine}
\usepackage{comment}

\usepackage{xcolor}
\usepackage[normalem]{ulem}
\usepackage{marginnote}

\ifdefined\showrevision

\newcommand{\revref}[2]{%
\marginnote{$R_{#1}C_{#2}$}
}

\usepackage{lipsum,multicol}
\usepackage{marginnote,xcolor}
\usepackage{changepage}
\usepackage{adjmulticol}

\newcommand{\mymarginnote}[4]{\marginnote{\begin{adjustwidth}{#1}{#2} 
\color{blue}            
$R_{#3}C_{#4}$
\end{adjustwidth}}
}

\newcommand{\revmod}[1]{%
{\color{blue}#1}
}

\newcommand{\revdel}[1]{%
{\color{darkgray}\sout{#1}}
}

\else

\newcommand{\mymarginnote}[4]{\ignorespaces}
\newcommand{\revref}[2]{\ignorespaces}
\newcommand{\revmod}[1]{#1}

\newcommand{\revdel}[1]{}

\fi

\usepackage{stmaryrd}
\usepackage{bbm}
\usepackage{breqn}
\usepackage{bm}
\usepackage{amsthm}
\usepackage{algorithmicx} 
\usepackage{algpseudocode} 
\usepackage{algorithm}

\usepackage{subfigure}

\usepackage{tikz}
\usetikzlibrary{backgrounds,automata}
\usetikzlibrary{bayesnet}
\usepackage{xspace}

\newcommand\Ttwo{\text{T}\textsubscript{2}\xspace}

\newcommand\Tonen{\text{T}\textsubscript{1}\xspace}
\newcommand\Tone{ce\text{T}\textsubscript{1}\xspace}
\newcommand\iUS{\text{iUS}\xspace}

\definecolor{LightGray}{gray}{0.95}
\definecolor{white}{rgb}{1.0, 1.0, 1.0}

\DeclareMathOperator{\KL}{KL}
\usepackage{accents}

\ifCLASSOPTIONcompsoc
  \usepackage[nocompress]{cite}
\else
  \usepackage{cite}
\fi
\ifCLASSINFOpdf
\else
\fi

\let\orighref\href
\renewcommand{\href}[2]{\orighref{#1}{#2\,\faExternalLink}}

\begin{document}
%

\title{Unified Cross-Modal \revmod{Medical} Image Synthesis with Hierarchical Mixture of Product-of-Experts}

\author{Reuben Dorent, Nazim Haouchine, Alexandra Golby, Sarah Frisken, Tina Kapur, William Wells

\IEEEcompsocitemizethanks{
\IEEEcompsocthanksitem R. Dorent, N. Haouchine, A. Golby, S. Frisken, T. Kapur and W. Wells are with Harvard Medical School and the Brigham and Women’s Hospital, MA, USA.
R. Dorent is also with MIND Team, Inria Saclay, Université Paris-Saclay, Palaiseau, France and Sorbonne Université, Institut du Cerveau - Paris Brain Institute - ICM, CNRS, Inria, Inserm, AP-HP, Hôpital de la Pitié Salpêtrière, F-75013, Paris, France. W. Wells is also with Massachusetts Institute of Technology, MA, USA.
}
}

\IEEEtitleabstractindextext{%
\begin{abstract}
We propose a deep mixture of multimodal
hierarchical variational auto-encoders called MMHVAE that synthesizes missing
images from observed images in different modalities.
MMHVAE’s design focuses on tackling four challenges: (i)
creating a complex latent representation of multimodal
data to generate high-resolution images; (ii) encouraging the
variational distributions to estimate the missing information
needed for cross-modal image synthesis; (iii) learning to
fuse multimodal information in the context of missing
data; (iv) leveraging dataset-level information to handle
incomplete data sets at training time. 
Extensive experiments are performed on the challenging problem of pre-operative brain multi-parametric magnetic resonance and intra-operative ultrasound imaging. 
\end{abstract}

}

\maketitle

\IEEEdisplaynontitleabstractindextext
\IEEEpeerreviewmaketitle


\section{Introduction}

\lettrine[findent=2pt]{\textbf{M}}{issing} data is a widespread problem in image analysis that presents significant practical and methodological challenges. This issue is common when the data collection is \textit{longitudinal} and \textit{multimodal}. For example, in medical image analysis,  different sensors (e.g., MRI, ultrasound imaging) collect patient data at various stages (diagnosis, monitoring, surgery, follow-up), often resulting in incomplete image sets~\cite{remind}. These challenges also arise  in dynamic environments like remote sensing~\cite{7284768} or automated driving~\cite{8943388}, where sensor availability and type may change over time. Consequently, handling missing data is essential for robust analysis in these multimodal scenarios.

The statistical literature offers an extensive set of data imputation methods to tackle missing data~\cite{little2019statistical}. These methods range from simple techniques, such as zero or mean imputation, to more advanced ones, like multiple imputation~\cite{1656662e-b94b-3fe2-9d22-0ff72b132245} and K-nearest neighbors (KNN)~\cite{10.1093/bioinformatics/17.6.520}. However, these methods primarily address missing data in tabular datasets and 
are typically not applicable to complex, high-dimensional data such as images. This limitation is particularly pronounced in the context of deep learning models since these models exhibit a strong sensitivity to subtle changes in the data distribution~\cite{Dong_2020_CVPR}. For these reasons, addressing missing data in image-based datasets requires techniques designed to tackle the complexity of images.

To tackle missing imaging data, deep-learning-based methods have been proposed to perform cross-modal image synthesis~\cite{wang2021review}, where missing data is estimated using observed data from other modalities. These methods include conditional generative adversarial networks (GANs)~\cite{isola2017image,park2019semantic,donnez2021realistic,9174648} and conditional variational auto-encoders\cite{chartsias2019disentangled}. Yet, these synthesis frameworks do not have the flexibility to handle missing data, requiring training for each possible combination of observed data.

To address these limitations, unified approaches have been proposed to perform inference from partially observed data. To have the flexibility for handling incomplete data, most approaches use zero imputation and are trained using artificially zero-imputed data~\cite{Sharma20,resvit,li2019diamondgan,lee2019collagan}.  Other methods bypass imputation, focusing instead on creating a common feature space for images across different modalities. This involves extracting and fusing modality-specific features through arithmetic operations like mean \cite{hemis,pimms,dorent2021learning}, max \cite{chartsias2017multimodal}, or a combination of sum, product and max \cite{zhou2020hi}. However, these methods often fail to encourage the network to learn a shared multimodal latent representation and need more theoretical foundations. A more principled solution is offered by multimodal Variational Auto-Encoders (MVAEs)~\cite{wu2018multimodal,dorent2019hetero}, which utilize a probabilistic fusion operation to create a shared representation space. However, the low-dimensional nature of the latent space in MVAEs typically leads to the generation of blurry synthetic images.  \revmod{In contrast, Hierarchical VAEs (HVAEs)~\cite{vahdat2020nvae,ranganath2016hierarchical,maaloe2019biva,sonderby2016ladder} have demonstrated improved generative performance by leveraging a more complex shared latent representation. Yet, HVAEs are designed for mono-modal settings and do not handle missing data. Another key common limitation across all these methods is the requirement for complete data during training, which often results in excluding a large part of the available data.

In this work, we introduce a hierarchical multimodal variational auto-encoder (MMHVAE) that performs cross-modal image synthesis and is designed to handle incomplete data at both training and inference times.}
The main contributions are summarized as follows:
\begin{enumerate}
    \item We propose a novel hierarchical multimodal variational auto-encoder to perform unified cross-modality image synthesis and allow for an incomplete set of images as input.
    \item We model the variational posterior as a mixture of Product-of-Experts. Each Product-of-Experts has a factorization similar to the true posterior. We demonstrate that this approach encourages the Product-of-Experts to not only encode observed information but also estimate missing information needed for image synthesis.
    \item In contrast to most existing approaches, we assume that training data may not always be complete. To regularize the distributions on the non-observed part of the training data, we introduce an adversarial strategy to leverage dataset-level information. 
    \item Intensive experiments are performed on the challenging problem of cross-modality image synthesis between multi-parametric Magnetic Resonance sequences and ultrasound data. This includes an evaluation of image synthesis using paired multimodal data and two relevant downstream tasks: image segmentation and image registration. Better image synthesis is obtained compared to existing unified cross-modality approaches while being less computationally heavy.

\end{enumerate}
This work is a substantial extension of our conference paper~\cite{dorent2023Unified}. Improvements include: 1) Generalization to more than two imaging modalities; 2) A novel method to leverage incomplete data at training time; 3) Additional mathematical proofs; 4) New experiments, including an ablation study and validation on two downstream tasks.

\begin{figure*}[tb!]
    \centering
    \subfigure[VAEs]{\includegraphics[width=0.18\textwidth]{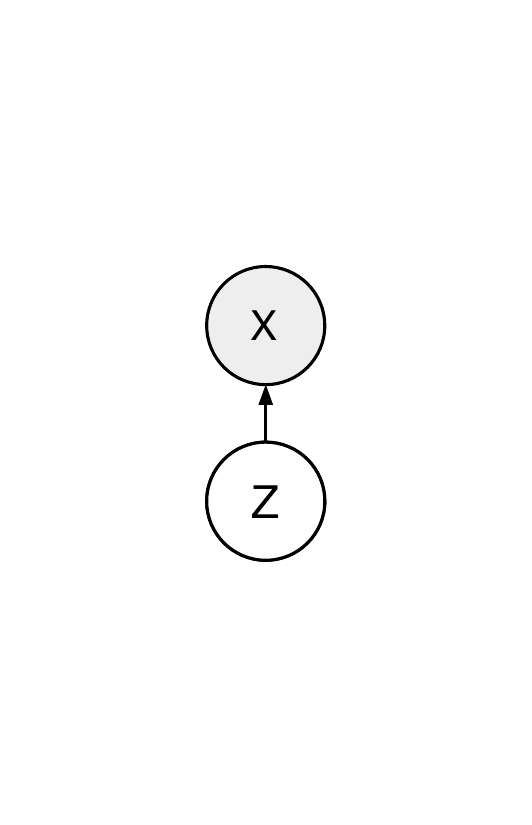}} 
    \subfigure[HVAEs]{\includegraphics[width=0.19\textwidth]{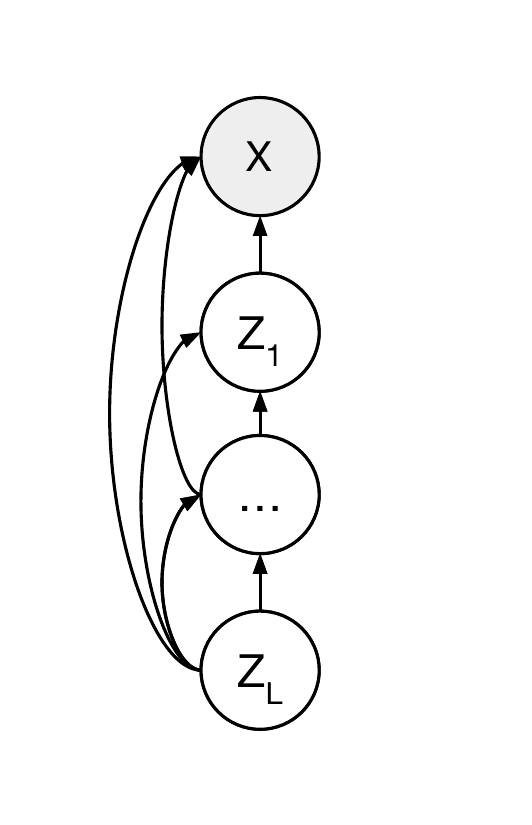}} 
    \subfigure[MVAEs]{\includegraphics[width=0.18\textwidth]{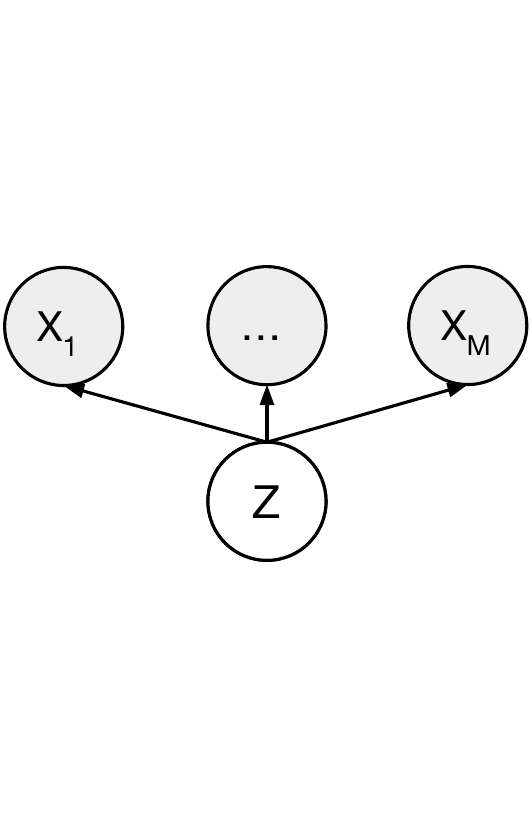}}
    \subfigure[MMHVAE (Ours)]{\includegraphics[width=0.18\textwidth]{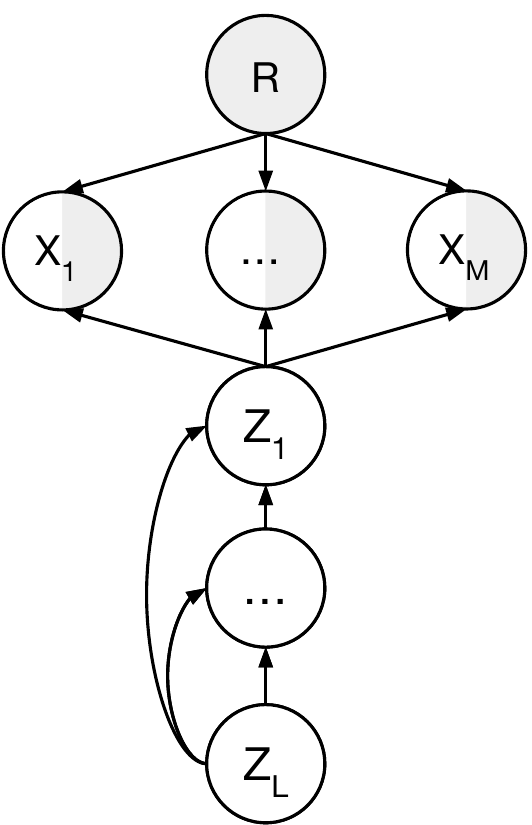}}
    \caption{Graphical models of: (a) variational auto-encoder (VAE); (b) hierarchical VAE (HVAE); (c) multimodal VAE (MVAE); (d) Our mixture of multimodal hierarchical VAE with missing data. Observed variables are in grey. In our model, variables are not always observed (partially gray). }
    \label{fig:graphical_models}
\end{figure*}

\section{Related Work}
In this section, we review Variational Auto-Encoders (VAEs), their hierarchical and multimodal extensions and state-of-the-art unified cross-modal synthesis frameworks.

\subsection{Variational Auto-Encoders (VAEs).}
The goal of Variational Auto-Encoders (VAEs)~\cite{kingma2013auto} is to train a generative model in the form of $p(\bm{x}, \bm{z})=p(\bm{z})p(\bm{x}|\bm{z})$ where $p(\bm{z})$ is a prior distribution (e.g. isotropic Normal distribution) over latent variables $\bm{Z}\in\mathbb{R}^{H}$  and where $p_{\theta}(\bm{x}|\bm{z})$ is a decoder parameterized by $\theta$ that generates data $\bm{X}\in\mathbb{R}^{\Omega}$ given $\bm{Z}$. Its graphical model is shown in Fig.\ref{fig:graphical_models}. 
The latent space dimension, $H$, is typically much lower than the image space dimension $\Omega$, i.e. $H\ll \Omega$. The training goal with respect to $\theta$ is to maximize the marginal likelihood of the data $p_{\theta}(\bm{x})$ (the ``evidence''); however since the true posterior $p_{\theta}(\bm{z}|\bm{x})$ is in general intractable, the variational evidence lower bound (ELBO) is instead optimized. The ELBO $\mathcal{L}_{\text{VAE}}(\bm{x}; \theta, \phi)$ is defined by introducing an approximate posterior $q_{\phi}(\bm{z}|\bm{x})$ with parameters $\phi$:
\begin{equation}
    \label{eqq:elbo_vae}
    \mathcal{L}_{\text{VAE}}(\bm{x}; \theta, \phi)  \triangleq \mathbb{E}_{q_{\phi}(\bm{z}|\bm{x})}[\log p_{\theta}(\bm{x}|\bm{z})] - \KL[q_{\phi}(\bm{z}|\bm{x})||p(\bm{z})] \ ,
\end{equation}
where $\KL[q||p]$ is the Kullback-Leibler divergence between distributions $q$ and $p$.

\subsection{Hierarchical Variational Auto-Encoders (HVAEs).} To increase the expressiveness of both the approximate posterior and prior, Hierarchical VAEs (HVAEs) have been introduced~\cite{vahdat2020nvae,ranganath2016hierarchical,maaloe2019biva,sonderby2016ladder}. In these extensions, the latent variable $\bm{Z}$ is partitioned  into disjoint groups, i.e. $\bm{Z}=\{Z_1,\dotsc,Z_L\}$, where $L$ is the number of groups. The dimension of the latent representations $Z_l$ typically exponentially decreases with the depth $l$. Then, the prior distribution is represented by \sloppy${p_{\theta}(\bm{z})=p(z_L)\prod_{l=1}^{L-1}p_{\theta}(z_l|\bm{z_{>l}})}$, where \sloppy${\bm{z_{>l}}=(z_k)_{k=l+1}^{L}}$, and the approximate posterior by \sloppy${q_{\phi}(\bm{z}|\bm{x})=q_{\phi}(z_L|\bm{x})\prod_{l=1}^{L-1}q_{\phi}(z_l|\bm{z_{>l}},\bm{x})}$. As HVAEs were originally designed to learn a mono-modal generative model,  they assume  that all the data $\bm{X}$ is presented at all times. 
In contrast, our approach has the flexibility to handle incomplete sets of multimodal images at training and testing times, enabling pratical cross-modal image synthesis.

\subsection{Multimodal Variational Auto-Encoders (MVAEs)}
Multimodal VAEs (MVAEs)~\cite{wu2018multimodal,dorent2019hetero,shi2019variational} extended VAEs by learning a multimodal data representation and supporting missing data at inference time. MVAEs assume that $M$ paired images \sloppy${\bm{X}=(X_1,\dotsc,X_M) \in \mathbb{R}^{M \times \Omega}}$ are conditionally independent given a shared representation $\bm{Z}$ as highlighted in Fig.\ref{fig:graphical_models}, i.e. $p_{\theta}(\bm{x}|\bm{z})=\prod_{i=1}^{M}p(x_i|\bm{z})$.

Instead of training one single variational  network $q_{\phi}(\bm{z}|\bm{x})$ that requires all images to be present at all times, MVAEs factorize the approximate posterior as a combination of $M$ unimodal variational distributions $(q_{\phi}(\bm{z}|x_j))_{j=1}^{M}$. Specifically, MVAE~\cite{wu2018multimodal} factorizes the variational posterior as a Product-of-Experts (PoE), i.e.: 
\begin{equation}
    q^{\text{PoE}}_{\phi}(\bm{z}|\bm{x})=p(\bm{z})\prod_{j=1}^{M}q_{\phi}(\bm{z}|x_j) \ .
\end{equation}

Alternatively, MMVAE~\cite{shi2019variational} proposed to decompose the variational posterior using a sum rule as a Mixture-of-Experts (MoE), i.e.: 
\begin{equation}
    q^{\text{MoE}}_{\phi}(\bm{z}|\bm{x})=\frac{1}{M}\sum_{j=1}^{M}q_{\phi}(\bm{z}|x_j) \ .
\end{equation}
While the sum rule has been shown to be most resilient to estimation errors~\cite{kittler1998combining}, the product rule follows a factorization similar to the true posterior~\cite{wu2018multimodal}.

Although the approximate posteriors of MVAEs have the flexibility to encode incomplete observations, these methods both face limitations in
performing cross-modal image synthesis.
The product rule typically generates misaligned latent representations across modalities, necessitating an ad hoc training sampling procedure to ensure similar representations for incomplete observations. In contrast, the sum rule focuses on generating all modalities from a single modality but does not combine representations when more than one modality is available.  \revmod{To address these limitations, MoPoE~\cite{sutter2021generalized} introduced a variational posterior modeled as a mixture of product-of-experts. However, similar to MVAE and MMVAE, it relies on low-dimensional latent representations, which are inadequate for high-resolution image synthesis. Moreover, these methods assume access to all modalities during training, limiting their ability to leverage incomplete observed datasets. To overcome these limitations, we propose in this work a model that (i) learns to effectively combine representations from multiple input modalities, (ii) accurately estimates missing information for cross-modal image synthesis, (iii) employs a more complex latent representation to enable high-resolution image synthesis, and (iv) leverages incomplete training data to learn more robust and generalizable representations.}

\subsection{Unified cross-modal image synthesis}
Several methodologies employing Generative Adversarial Networks (GANs)~\cite{lee2019collagan,li2019diamondgan,Sharma20} and Transformers~\cite{resvit} have been proposed for unified cross-modal image synthesis, where a single model is utilized to synthesize all imaging modalities given any combination of incomplete observed data. However, these approaches omit the key aspect of learning a shared representation among multimodal data. Typically, these techniques concatenate multimodal images and substitute the missing images with zero tensors, i.e., zero-imputation. During training, all imaging data is assumed to be available. More specifically, the training procedure involves randomly substituting images with zero tensors and learning to synthesize the substituted images using supervised learning. \revmod{More recently, cross-modal image synthesis frameworks based on diffusion models have been proposed. For instance, Cola-DIFF~\cite{jiang2023cola} employs a latent diffusion model specifically designed for brain MRI, but it only supports a many-to-one synthesis setting, making it unsuitable for unified synthesis. In contrast, some unified diffusion-based approaches have been introduced, where missing modalities are initialized with random noise and iteratively refined in the image space~\cite{meng2022novel,meng2024multi,xiao2024fgc2f}. However, these methods do not learn compact latent representations and typically require large training datasets to prevent memorization~\cite{gu2023memorization}. They are also computationally intensive at inference time.  Finally, like other unified approaches based on GANs or Transformers, they assume access to complete training data.
In contrast to all these methods, our framework is designed to learn a common representation of multimodal data and has the flexibility to leverage incomplete training sets.}

\section{Mixture of Multimodal Hierarchical Variational Auto-Encoders}
In this paper, we propose a deep mixture of multimodal hierarchical VAE called MMHVAE that synthesizes missing modalities from observed images in different modalities. MMHVAE’s design focuses on tackling four challenges: (i) creating a complex latent representation of multimodal data to generate high-resolution images; (ii) encouraging the variational distributions to estimate the missing information needed for cross-modal image synthesis; (iii) learning to fuse multimodal information in the context of missing data; (iv) leveraging dataset-level information to handle incomplete data sets at training time.

\subsection{Hierarchical latent 
representation of multimodal images}\label{sec:model_hiearachy} 
Let the random variable $\bm{X}=(X_1,\dotsc,X_M) \in \mathbb{R}^{M \times \Omega}$ be a complete set of paired (i.e. co-registered) multimodal images, where $M$ is the total number of image modalities and $\Omega$ the number of pixels. The images $\bm{X}$ are assumed to be conditionally independent given a latent random variable $\bm{Z}$. Then, the conditional distribution $p_{\theta}(\bm{x}|\bm{z})$ parameterized by $\theta$ can be written as:
\begin{equation}
    p_{\theta}(\bm{x}|\bm{z})=\prod_{j=1}^{M}p_{\theta}(x_j|\bm{z}) \ .
    \label{eq:graphicalmodel}
\end{equation}

Given that VAEs and MVAEs typically produce blurry images, we propose to use a hierarchical representation of the latent variable $\bm{Z}$ to increase the expressiveness of the model as in HVAEs. Specifically, the latent variable $Z$ is partitioned into disjoint groups, as shown in Fig.\ref{fig:graphical_models} i.e. $\bm{Z}=(Z_1,\dotsc, Z_L)$, where $L$ is the number of groups. The prior $p(\bm{z})$ is  then represented by:
\begin{equation}
    p_{\theta}(\bm{z})=p(z_L)\prod_{l=1}^{L-1}p_{\theta_{l}}(z_l|\bm{z_{>l}}) \ ,
\end{equation}
where $\bm{z_{>l}}=(z_k)^{L}_{k=l+1}$, $p(z_L)=\mathcal{N}(z_L; \bm{0}_{H_L}, I_{H_L})$ is an isotropic Normal prior distribution and the conditional prior distributions $p_{\theta_{l}}(z_l|\bm{z_{>l}})$ are  Normal distributions with mean and diagonal covariance parameterized using neural networks, i.e. $p_{\theta_{l}}(z_l|\bm{z_{>l}})=\mathcal{N}(z_l; \mu_{\theta_{l}}(\bm{z_{>l}}), D_{\theta_{l}}(\bm{z_{>l}}))$. Note that the dimension of the latent representations exponentially decreases with the depth. In particular, the coarsest latent representation is a global descriptor of the multimodal images of size $H_L$ (e.g. $H_L=128$), while the finest latent variable representation corresponds to a set of multimodal pixel descriptors of size $F$, i.e. $H_1=\Omega\times F$. Thanks to the hierarchy, we can learn a complex prior for the high-dimensional representation $z_1$ with complex local correlations characteristic of real images. In this work, we assume that the finest latent representation $z_1$ carries all the information required to describe each image, i.e.:
\begin{equation}
    p_{\theta}(\bm{x}|\bm{z})=\prod_{j=1}^{M}p_{\theta_{j}}(x_j|z_1) \ ,
\end{equation}
where the image decoding distributions are modeled as Normal with fixed variance $\sigma$, i.e. \sloppy${p_{\theta_{j}}(x_j|z_1)=\mathcal{N}(x_j;\mu_{\theta_j}(z_1), \sigma I)}$.

\subsection{Marginal log-likelihood objective with incomplete data}\label{sec:methods_objective}
In this work, we face the problem of missing data at inference and training time. We denote the vector of missingness indicators $\bm{R}$ with $R_{j}=1$ if $X_j$ is observed and $0$ otherwise. The partially observed samples $\bm{X}$ can be divided into an observed component $\bm{X}^o_{\bm{R}}=\{\bm{X}_i, \text{ such that } R_i=1\}$ and a missing component $\bm{X}^m_{\bm{R}}=\{\bm{X}_i, \text{ such that } R_i=0\}$. 

Let $\mathcal{X}\times\mathcal{R}=\{(\bm{x}_1,\bm{r}_1)),..,(\bm{x}_N,\bm{r}_N)\}$ be a training set of $N$ partially observed images and $\mathcal{X}^{o}=\{ \bm{x}^{o}_{1; \bm{r}_1},..,\bm{x}^{o}_{N; \bm{r}_N}\}$ its restriction to the observed component.

The overall objective is to maximize the expected observed marginal log-likelihood, i.e. $\mathbb{E}_{(\bm{x},\bm{r})\sim p_{\text{data}}}\left[\log p_{\theta}(\bm{x}^{o}_{\bm{r}})\right]$. As the true posterior is intractable, the expected value of a tractable lower-bound is instead maximized. While the expected value is estimated using the samples in $\mathcal{X}^{o}$,  a variational distribution $q_{\phi}(\bm{z}|\bm{x}^{o}_{\bm{r}})$ is introduced to approximate the posterior  $p_{\theta}(\bm{z}|\bm{x}^{o}_{\bm{r}})$. 

\subsection{Modeling the variational posterior as a mixture}
In this work, we propose to model the variational posterior as a mixture distribution, where each component approximates the true posterior with incomplete input data. We show that this encourages the mixture components to encode all available information and estimate the missing information needed for cross-modal image synthesis.

Let $S$ be the set of all $2^M-1$ vectors of length $M$ consisting solely of zeros and ones, with each vector having at least one non-zero entry. For any vector $\bm{r}\in S$, let $S_{\bm{{r}}}$ be the subset of indicators in $S$ that share the zero indices of $\bm{r}$, i.e. \sloppy${S_{\bm{{r}}}=\{\bm{r'}\in S, \text{ s.t.: } \ r'_i=0 \text{ if } r_i=0\}}$ (e.g. \sloppy${S_{{{[1,1,0,0]}}}=\{[1,1,0,0], [1,0,0,0], [0,1,0,0] \}}$). 

We propose to express the variational $q_{\phi}(\bm{z}|\bm{x}^{o}_{\bm{r}})$  as a mixture (convex combination) of distributions represented by:
\begin{equation}
    q^{\text{MMHVAE}}_{\phi}(\bm{z}|\bm{x}^{o}_{\bm{r}}) = \sum_{\bm{r'}\in S_{\bm{{r}}}} \alpha^{(\bm{r})}_{\bm{r'}} q_{\phi}(\bm{z}|\bm{x}^{o}_{\bm{r'}}) \,
\end{equation}
where the input data $\bm{x}^{o}_{\bm{r'}}$ is a subset of the observations $\bm{x}^o_{\bm{r}}$ and $\alpha^{(\bm{r})}_{\bm{r'}}$ are the mixture weights such that $\alpha^{(\bm{r})}_{\bm{r'}}\geq 0$  and $\sum_{\bm{r'}\in S_{\bm{{r}}}} \alpha^{(\bm{r})}_{\bm{r'}}=1$. These weights are hyper-parameters.

Let  $p_{\alpha}(\cdot|\bm{R}=\bm{r})$ denote the distribution of the mixture components $\bm{R'}$ given $\bm{R}$, i.e. $p_{\alpha}(\bm{r'}|\bm{r})=\alpha^{(\bm{r})}_{\bm{r'}}$. We can demonstrate that the evidence $\log p(\bm{x}^{o}_{\bm{r}})$ is lower-bounded by a new ELBO $\mathcal{L}^{\text{ELBO}}_{\text{MMHVAE}}(\bm{x}^{o}_{\bm{r}}; \theta, \phi)$ defined as follows:
\begin{equation}
\label{eq:elbodefMMHVAE}
\log p(\bm{x}^{o}_{\bm{r}}) \geq  \mathcal{L}^{\text{ELBO}}_{\text{MMHVAE}}(\bm{x}^{o}_{\bm{r}}; \theta, \phi) \triangleq  \mathbb{E}_{\bm{r'}\sim p_{\alpha}(\cdot|\bm{r})} \left[ \mathcal{L}(\bm{x}^{o}_{\bm{r}}, \bm{r'}; \theta, \phi) \right]
\end{equation}

\noindent where:

\begin{equation}
\begin{split}
\label{eq:elbo_MMHVAE_all}
\mathcal{L}(\bm{x}^{o}_{\bm{r}}, \bm{r'}; \theta, \phi) &=  \sum_{\substack{j=1 \\ \text{s.t. } r'_j=1}}^{M}\underbrace{\mathbb{E}_{q_{\phi}(\bm{z}|\bm{x}^{o}_{\bm{r'}})}[\log p_{\theta}(x_j|\bm{z})]}_{\text{reconstruction input $x_j$}} \\
& \quad + \sum_{\substack{j=1 \\ \text{s.t. } (1-r'_j)r_j=1}}^{M}\underbrace{\mathbb{E}_{q_{\phi}(\bm{z}|\bm{x}^{o}_{\bm{r'}})}\left[\log p_{\theta}(x_j|\bm{z})\right]}_{\text{synthesis of $x_j$ using  $\bm{x}^{o}_{\bm{r'}}$}} \\
& \quad - \underbrace{\KL\left[q_{\phi}(\bm{z}|\bm{x}^{o}_{\bm{r'}})\Vert p_{\theta}(\bm{z})\right]}_{\text{regularization}}  \ .
\end{split}
\end{equation}

\begin{proof} See Appendix 1.1.
\end{proof}

The proposed ELBO $\mathcal{L}^{\text{ELBO}}_{\text{MMHVAE}}(\bm{x}^{o}_{\bm{r}}; \theta, \phi)$ is the expected value of $\mathcal{L}(\bm{x}^{o}_{\bm{r}}, \bm{R'}; \theta, \phi)$, which contains input reconstruction, cross-modal image synthesis, and regularization terms.

This formulation not only allows optimization of a tractable upper-bound of the log likelihood, it also encourages the variational posterior to estimate all the information required to generate all the images. As demonstrated in Appendix 1.2, the ELBO $\mathcal{L}^{\text{ELBO}}_{\text{MMHVAE}}(\bm{x}^{o}_{\bm{r}}; \theta, \phi)$ is in fact maximal with respect to the variational parameters $\phi$ when the variational components $q_{\phi}(\bm{z}|\bm{x}^{o}_{\bm{r'}})$ are equal to the true posterior distribution $p_{\theta}(\bm{z}|\bm{x}^o_{\bm{r}})$. Consequently, the objective of the variational distribution $q_{\phi}(\bm{z}|\bm{x}^{o}_{\bm{r'}})$ is to encode the information needed to generate the input data ($\{ x_i, \text{ s.t. } r'_i=1 \}$) while estimating the missing information to generate the non-encoded ones ($\{ x_i, \text{ s.t. } r'_i=0 \text{ and } r_i=1 \}$). This especially encourages the latent representations to be aligned across any subset of modalities.

\subsection{Variational parameterization for fusing incomplete multimodal inputs}\label{sec:parametrization}
Our next objective is to construct a variational component for each non-empty subset of observed data that approximates the posterior distribution. A straightforward approach would involve creating a variational distribution for every possible set of input data, resulting in the handling of $2^M-1$ encoding networks. Inspired by MVAEs~\cite{wu2018multimodal,shi2019variational}, we instead propose to create $M$ unimodal encoding networks. This section demonstrates that we can identify candidates for the variational components that can be expressed (i) with a factorization similar to the true posterior $p_{\theta}(\bm{z}|\bm{x}^{o}_{\bm{r}})$; (ii) as a combination of variational unimodal distributions; and (iii) by fusing input information using a principled operation at each level of the hierarchy.

\begin{figure*}[tb!]
    \centering
    \includegraphics[width=0.90\textwidth]{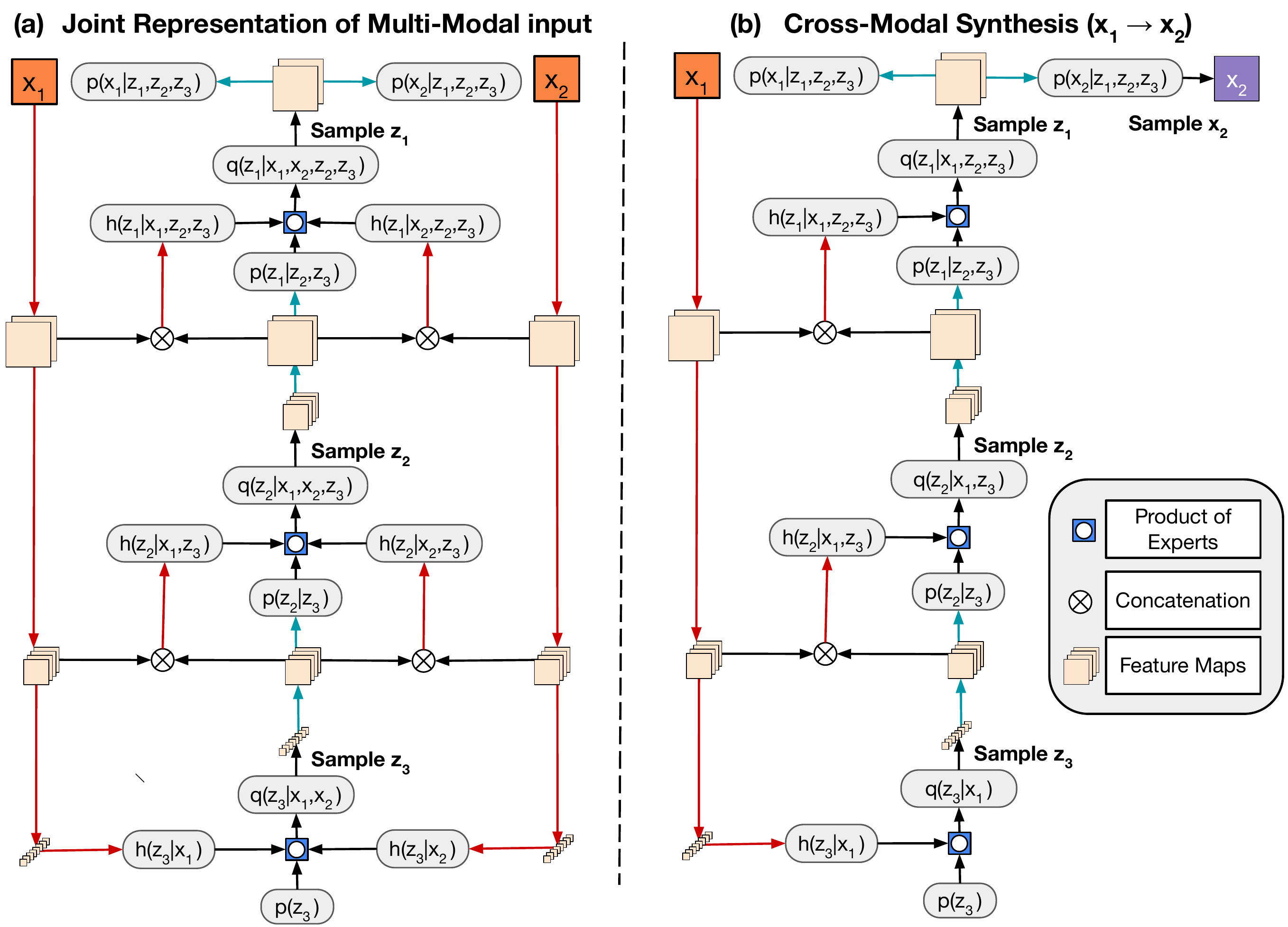}
    \caption{The neural networks implementing (a) the encoder $q(\bm{z}|\bm{x})$ and decoder $p(\bm{x}|z_1)$; (b) the encoder $q(\bm{z}|x_1)$ and decoder $p(\bm{x}|z_1)$ for a $L=3$ group hierarchical VAE with $M=2$ modalities. }
    \label{fig:model}
\end{figure*}

Let $\bm{x}^o_{\bm{r}}$ be a set of observed images. The true hierarchical posterior $p_{\theta}(\bm{z}|\bm{x}^o_{\bm{r}})$ can be factorized as a hierarchical combination of the conditional unimodal distributions $((p_{\theta}(z_l|x_{j},\bm{z_{>l}}))_{j=1  \text{ s.t. } r'_j=1}^{M})_{l=1}^{L}$ and the prior distribution $(p_{\theta_l}(z_l|\bm{z_{>l}}))_{l=1}^{L}$ as follows:

\begin{multline}
    \label{eq:trueposterior_factorization}
    p_{\theta}(\bm{z}|\bm{x}^o_{\bm{r}})\propto  \underbrace{\left(p(z_L)\prod_{\substack{j=1 \\ \text{s.t. } r_j=1}}^{M}\frac{\overbrace{p_{\theta}(z_L|x_j)}^{\text{unimodal}}}{p(z_L)}\right)}_{\propto p_{\theta}(z_L|\bm{x}^o_{\bm{r}})} \\
    \prod_{l=1}^{L-1}\underbrace{\left(p_{\theta_l}(z_l|\bm{z_{>l}})\prod_{\substack{j=1 \\ \text{s.t. } r_j=1}}^{M}  \frac{\overbrace{p_{\theta}(z_l|x_j,\bm{z_{>l}})}^{\text{unimodal}}}{p_{\theta_l}(z_l|\bm{z_{>l}})}\right)}_{\propto p_{\theta}(z_l|\bm{x}^o_{\bm{r}},\bm{z_{>l}})} \ .
\end{multline}

\begin{proof}  See Appendix 1.3.
\end{proof}
Unlike the conditional prior distributions $p_{\theta_l}(z_l|\bm{z_{>l}})$ parameterized using using neural networks in~\ref{sec:model_hiearachy}, the conditional unimodal distributions $p_{\theta}(z_l|x_j,\bm{z_{>l}})$ are intractable. We thus propose a variational approach. Specifically, let $\bm{r'}$ be an indicator vector in $S_{\bm{{r}}}$. We propose to factorize the variational posterior distribution $q(\bm{z}|\bm{x}^{o}_{\bm{r'}})$ similarly to the true posterior $p_{\theta}(\bm{z}|\bm{x}^o_{\bm{r}})$  defined in \eqref{eq:trueposterior_factorization} as a product of conditional prior distributions $p_{\theta_l}(z_l|\bm{z_{>l}})$ and variational unimodal distributions $(q(z_l|x_j,\bm{z_{>l}}))_{j\in\{1,..,M\} \text{ s.t. } r'_j=1}$:

\begin{multline}
    \label{eq:variational_factor}
    q(\bm{z}|\bm{x}^{o}_{\bm{r'}})\propto \underbrace{\left( p(z_L)\prod_{\substack{j=1 \\ \text{s.t. } r'_j=1}}^{M}\frac{q(z_L|x_j)}{p(z_L)}\right)}_{q(z_L|\bm{x}^o_{\bm{r'}})} \\
    \prod_{l=1}^{L-1}\underbrace{\left(p_{{\theta}_{l}}(z_l|\bm{z_{>l}})\prod_{\substack{j=1 \\ \text{s.t. } r'_j=1}}^{M} \frac{q(z_l|x_j,\bm{z_{>l}})}{p_{\theta_{l}}(z_l|\bm{z_{>l}})}\right)}_{q_{\phi_l,\theta_l}(z_l|\bm{x}^o_{\bm{r'}},\bm{z_{>l}})} \ .
\end{multline}

While the product of two Normal densities $p_1$ and $p_2$ is proportional to a Normal density with closed form solutions for its parameters, the ratio $\frac{p_1}{p_2}$ is a Normal density if and only if the $p_1$'s variance is element-wise larger than  $p_2$'s variance. To satisfy this constraint, we propose a parametrization of the approximate conditional unimodal distributions $q(z_l|x_j,\bm{z_{>l}})$ that guarantees $\frac{q(z_l|x_j,\bm{z_{>l}})}{p_{\theta_l}(z_l|\bm{z_{>l}})}$ to always be a Normal distribution. Specifically, the approximate conditional unimodal distributions $q(z_l|x_j,\bm{z_{>l}})$ are defined as a product of the conditional prior $p_{\theta_l}(z_l|\bm{z_{>l}})$ and residual conditional unimodal Normal distributions $h_{\phi^{j}_{l}}(z_l|x_j,\bm{z_{>l}})$:
\begin{equation}
q(z_l|x_j,\bm{z_{>l}}) = p_{\theta_l}(z_l|\bm{z_{>l}}) h_{\phi^{j}_{l}}(z_l|x_j,\bm{z_{>l}}) \ ,
\end{equation}
where $h_{\phi^{j}_{l}}(z_l|x_j,\bm{z_{>l}})$ is a factorized Normal distribution parameterized using neural networks, i.e. \sloppy ${h_{\phi^{j}_{l}}(z_l|x_j,\bm{z_{>l}})=\mathcal{N}(z_l; \mu_{\phi^{j}_{l}}(x_j,\bm{z_{>l}}); D_{\phi^{j}_{l}}(x_j,\bm{z_{>l}}))}$.

Consequently, the expression of the variational distribution $q(\bm{z}|\bm{x}^{o}_{\bm{r'}})$ in \eqref{eq:variational_factor} can be simplified as:
\begin{multline}
\label{eq:variational_factor_simplified}
    q_{\phi,\theta}(\bm{z}|\bm{x}^{o}_{\bm{r'}})
    \propto \underbrace{\left( p(z_L)\prod_{\substack{j=1 \\ \text{s.t. } r'_j=1}}^{M}h_{\phi_{L}}(z_L|x_j)\right)}_{q_{\phi_{L},\theta_{L}}(z_L|\bm{x}^{o}_{\bm{r'}})} \\
    \prod_{l=1}^{L-1}\underbrace{\left(p_{\theta_l}(z_l|\bm{z_{>l}})\prod_{\substack{j=1 \\ \text{s.t. } r'_j=1}}^{M} h_{\phi^{j}_{l}}(z_l|x_j,\bm{z_{>l}})\right)}_{q_{\phi_{l},\theta_{l}}(z_l|\bm{x}^{o}_{\bm{r'}},\bm{z_{>l}})} \ .
\end{multline}
The factorization in \eqref{eq:variational_factor_simplified} shows that we can express all the $2^M-1$ variational posteriors using only $M$ unimodal variational distributions. Importantly, a closed form solution allows merging unimodal contributions from each available modality at each level $l\in\{1,\dotsc,L-1\}$ of the hierarchy. Specifically,  the conditional variational distribution $q_{\phi_l,\theta_l}(z_l|\bm{x}^o_{\bm{r'}},\bm{z_{>l}})$ is a Normal distribution with mean $\mu_{\phi_{l},\theta_{l}}(\bm{x}^o_{\bm{r'}},\bm{z_{>l}})$ and diagonal covariance $D_{\phi_{l},\theta_{l}}(\bm{x}^o_{\bm{r'}},\bm{z_{>l}})$ defined by:
\begin{equation}
    \begin{cases}
        D_{\phi_{l},\theta_{l}}(\bm{x}^o_{\bm{r'}},\bm{z_{>l}})=\left(D_{\theta_{l}}(\bm{z_{>l}})^{-1} + \sum\limits_{\substack{j=1 \\ \text{s.t.} \\ r'_j=1}}^M D_{\phi^{j}_{l}}(x_{j},\bm{z_{>l}})^{-1}\right)^{-1} \\
        \\
        \mu_{\phi_{l},\theta_{l}}(\bm{x}^o_{\bm{r'}},\bm{z_{>l}}) = D_{\phi_{l},\theta_{l}}(\bm{x}^o_{\bm{r'}},\bm{z_{>l}})^{-1} \\
         \left(
    D_{\theta_{l}}(\bm{z_{>l}})^{-1}\mu_{\theta_{l}}(\bm{z_{>l}}) 
    + \sum\limits_{\substack{j=1 \\ \text{s.t.} \\ r'_j=1}}^M D_{\phi^{j}_{l}}(x_{j},\bm{z_{>l}})^{-1}\mu_{\phi^{j}_{l}}(x_{j},\bm{z_{>l}}) \right) 
    \end{cases}
    \label{product_gaussian_formula}
\end{equation} 
with $D_{\theta_{L}}(z_L)=I_{H_L}$ and $\mu_{\theta_{L}}(z_L)=\bm{0}_{H_L}$

\begin{algorithm}
\flushleft
  \caption{Training algorithm for MMHVAE. $B$ is the batch size. $T$ is the number of random subsets of observed images encoded at each training iteration. }
  \label{alg:training}
  \begin{algorithmic}[1] 
      \State $\theta, \phi, \psi \gets \text{Initialize parameters}$   
    \Repeat
    \State $\bm{r} \gets$ Random indicator (drawn from $\mathcal{R}$ dataset)
    \State $\bm{x_{r}}^{o} \gets$ Random minibatch of $B$ multimodal images associated with indicator $\bm{r}$ (drawn from $\mathcal{X}$ dataset)
    \Statex \hspace{\algorithmicindent} \textit{\textbf{Data model optimization}}
    \State $\mathcal{L}^{KL} \gets 0$   
    \State $\mathcal{L}^{Img} \gets 0$  
    \State $\mathcal{L}^{GAN} \gets 0$  
    \For{$k \gets 1$ to $T$} \Comment{Samples from the mixture}
        \State $\bm{r'} \gets$ Random indicator drawn from $p_{\alpha}(\cdot|\bm{r})$
        \For{$l \gets L$ to $1$} \Comment{Encoding phase}
            \State $z_l \gets$ Random sample from $q_{\phi_{l},\theta_{l}}(z_l|\bm{x}^{o}_{\bm{r'}},\bm{z_{>l}})$
            \State $\mathcal{L}^{KL}_{l} \gets \KL \left[ q_{\phi_{l},\theta_{l}}(z_l|\bm{x}^{o}_{\bm{r'}},\bm{z_{>l}})\Vert p_{\theta_{l}}(z_l|\bm{z_{>l}})\right]$
            \State $\mathcal{L}^{KL} \gets \mathcal{L}^{KL} + \mathcal{L}^{KL}_{l} $  
        \EndFor
        \For{$j \gets 1$ to $M$} \Comment{Decoding phase}
            \State $\hat{x_j} \gets \mu_{\theta^{x}_j}(z_1)$
            \State $\mathcal{L}^{GAN} \gets  \mathcal{L}^{GAN} + d^{j}_{\psi_j}(\hat{x_j})$
            \If{$r_j=1$}
                \State $\mathcal{L}^{Img} \gets  \mathcal{L}^{Img} + \Vert x_j - \hat{x_j} \Vert^2 $
            \EndIf
        \EndFor    
    \EndFor
    \State $\mathcal{L}^{Total} \gets \mathcal{L}^{Img} +\lambda_{GAN}\mathcal{L}^{GAN} + \lambda_{KL}\mathcal{L}^{KL}$
    \State $\bm{g} \gets \nabla_{\theta,\phi}\mathcal{L}^{Total}$
    \State $\theta, \phi \gets$ Update parameters using gradient $\bm{g}$  
    \Statex \hspace{\algorithmicindent} \textit{\textbf{Discriminators optimization}}
    \For{$j \gets 1$ to $M$} 
        \If{$r_j=1$}
            \State $m_j \gets x_j$ \Comment{Memory for future missing $x_j$}
        \EndIf
        \State $\mathcal{L}^{Adv} \gets d^{j}_{\psi_j}(m_j) - d^{j}_{\psi_j}(\hat{x_j}) $
        
        \State $\bm{g} \gets \nabla_{\psi_{j}}\mathcal{L}^{Adv}$
        \State $\psi_{j} \gets$ Update parameters using gradient $\bm{g}$  
    \EndFor       
    \Until{ convergence of parameters $(\theta,\phi,\psi)$}
\end{algorithmic}
\end{algorithm}

\subsection{Regularizing image distributions using GAN loss}\label{sec:methods_regularization}
When maximizing the expected log-likelihood, our objective is to identify optimal parameter values that make the observed data most probable. However, the unobserved component of the data distribution $p_{\theta}(\bm{X_{R}}^{m})$ is disregarded, potentially resulting in out-of-distribution samples for the unobserved part of the data. To mitigate this issue, we propose to regularize the data distribution $p_{\theta}(\bm{X})$ through adversarial learning. Specifically, for each modality $j$, a discriminator network $d^{j}_{\psi{j}}$, parameterized by weights $\psi_{j}$, is trained to differentiate between real $j$ images from $\mathcal{X}^{o}$ and $j$ samples drawn from $p_{\theta}$, leading to the following minimization objective:
\begin{multline}
    \min_{\psi_{j}} \
    \mathbb{E}_{x_{j}}\left[ d^{j}_{\psi_{j}} \left(x_j \right) \right] 
    - \mathbb{E}_{\bm{x}^{o}_{\bm{r'}}}\left[ \mathbb{E}_{q_{\phi}(\bm{z}|\bm{x}^{o}_{\bm{r'}})}\left[d^{j}_{\psi_{j}}\left(\mu_{\theta^{x}_j}\left(z_1\right)\right)\right] \right] 
\end{multline}

Simultaneously, the data distribution model is trained to confound the discriminators $(d^{j}_{\psi{j}})_{j=1}^{M}$. Consequently, an additional GAN term $\mathcal{L}^{\text{GAN}}(\bm{r'};\theta,\phi)$ is introduced:
\begin{equation}
\label{eq:gan_term}
    \mathcal{L}^{\text{GAN}}(\bm{r'}; \theta,\phi) = \sum\limits_{j=1}^M 
    \mathbb{E}_{q_{\phi}(\bm{z}|\bm{x}^{o}_{\bm{r'}})}\left[d^{j}_{\psi_{j}}\left(\mu_{\theta^{x}_j}\left(z_1\right)\right)\right]
\end{equation}
In particular, this GAN term encourages samples from the non-observed data distribution to be realistic by leveraging modality-specific knowledge at the dataset level.

\subsection{Training strategy}
Combining sections \ref{sec:methods_objective} and \ref{sec:methods_regularization}, the complete training objective is to maximize the expected log-likelihood using the expected lower-bound \eqref{eq:elbo_MMHVAE_all} while minimizing the expected regularization term \eqref{eq:gan_term}, leading to the following minimization objective:
\begin{equation}
    \min_{\theta,\phi}\ \mathbb{E}_{\bm{r}\sim p_{data}}\mathbb{E}_{\substack{\bm{x}\sim p_{data}(\cdot|\bm{r}) \\ \bm{r'}\sim p_{\alpha}(\cdot|\bm{r})}}\left[ \mathcal{L}^{Total}(\bm{x}^{o}_{\bm{r}}, \bm{r'}; \theta, \phi)\right]
\end{equation}
where $\mathcal{L}^{Total}$ is the sum the negative ELBO defined in \eqref{eq:elbo_MMHVAE_all} and the regularization term in \eqref{eq:gan_term}:
\begin{equation}
\mathcal{L}^{Total}(\bm{x}^{o}_{\bm{r}}, \bm{r'}; \theta, \phi) = - \mathcal{L}(\bm{x}^{o}_{\bm{r}}, \bm{r'}; \theta, \phi) + \lambda_{GAN}\mathcal{L}^{GAN}(\bm{r'}; \theta, \phi) 
\end{equation}

We use a stochastic gradient descent approach to minimize the expected loss $\mathcal{L}^{Total}$. First, we draw a random indicator $\bm{r}$ from $\mathcal{R}$. Second, we draw a random mini-batch of size $B$ of observed images from $\mathcal{X}^{o}$ associated with the indicator $\bm{r}$. Third, we draw $T$ vectors $\bm{r'}\in S_{\bm{{r}}}$ from $p_{\alpha}(\cdot|\bm{r})$. Finally, we minimize the average loss of $\mathcal{L}^{Total}(\bm{x}^{o}_{\bm{r}}, \bm{r'}; \theta, \phi)$ for all drawn multimodal images $\bm{x}^{o}_{\rm{r}}$ in the mini-batch and all drawn vectors $\bm{r'}$. Note that the expected values in $\mathcal{L}^{Total}$ are estimated using a unique sample at each level of the hierarchy as in~\cite{maaloe2019biva}. Consequently, given a triplet $(\bm{r},\bm{x}^{o}, \bm{r'})$, the total loss $\mathcal{L}^{Total}(\bm{x}^{o}_{\bm{r}}, \bm{r'}; \theta, \phi)$ is defined as:
\begin{equation}
\begin{split}
    \mathcal{L}^{Total}(\bm{x}^{o}_{\bm{r}}, \bm{r'}; \theta, \phi) =& \sum_{j=1}^{M}||\mu_{\theta^x_j}(z_1)-x_j||^2  \\
    & + \lambda_{KL}\sum_{l=1}^{L}\log \frac{q_{\phi_{l},\theta_{l}}(z_l|\bm{x}^{o}_{\bm{r'}},\bm{z_{>l}})}{p_{\theta_{l}}(z_l|\bm{z_{>l}})} \\
    & +\lambda_{GAN}\sum_{j=1}^{M} d^{j}_{\psi_j}\left( \mu_{\theta^x_j}(z_1) \right) \ ,
\end{split}
\end{equation}
where $\lambda_{KL}$ and $\lambda_{GAN}$ weight the contributions of the KL divergence and GAN loss.  The complete training strategy is presented in Algorithm \ref{alg:training}. 

\section{Experimental setup}
To assess the performance of our model, we conducted comprehensive experiments on the challenging task of cross-modal synthesis of multi-parametric magnetic resonance images (MRI) and intraoperative ultrasound images (\iUS) in patients with brain tumors. Specifically, our focus was on the key MR sequences for brain tumor surgery, namely, the contrast-enhanced \Tonen-weighted (\Tone), \Ttwo-weighted, and \Ttwo Fluid Attenuation Inversion Recovery (FLAIR) sequences, resulting in $M=4$ modalities. MMHVAE is empirically evaluated by (i) cross-modal image synthesis results with standard paired evaluation metrics, (ii) an ablation study analyzing each novel component, (iii) segmentation results as a downstream task, and (iv) image registration results as a downstream task.

\subsection{Datasets}
Three datasets were utilized in our experiments:

\textbf{1. ReMIND:} The Brain Resection Multimodal Imaging Database (ReMIND) comprises pre-operative multi-parametric and intra-operative ultrasound data collected from 114 consecutive patients~\cite{remind}. For this study, we selected all the patients ($N=104$) who underwent acquisition of both pre-operative 3D MRI and intraoperative pre-dural opening ultrasound (iUS), reconstructed from a tracked handheld 2D probe. Not all 3D MR sequences were acquired for each patient, resulting in 98 \Tone, 66 \Ttwo, and 21 FLAIR MR scans. In particular, only 13 patients have a complete set of MRI, highlighting the need for methods designed to handle missing data. 
To create a paired dataset, pre-operative MR images were affinely registered with the pre-dura \iUS using NiftyReg~\cite{MODAT2010278}, following the pipeline described in~\cite{drobny2018registration}. Three neurological experts manually checked the registration outputs. Images were resampled to an isotropic $0.5\text{mm}$ resolution and padded for an in-plane matrix of $(192,192)$. While the ultrasound images were acquired with the same ultrasound probe, preoperative MR images were collected under various clinical protocols and scanners at multiple institutions. This led to different intensity distributions for a given MR sequence.

\textbf{2. UPenn-GBM:} The UPenn-GBM dataset comprises multi-parametric preoperative MRI data for brain tumor patients with delineations of the distinct tumor sub-regions~\cite{bakas2022university}. Ultrasound images are not available in this dataset. In contrast, all MR sequences were acquired pre-operatively for $N=611$ patients. All images are co-registered to the same anatomical template and resampled to an isotropic $1\text{mm}$ resolution. The brain images in UPenn-GBM are also included in the BraTS dataset~\cite{bakas2019identifying}, albeit without undergoing skull-stripping preprocessing. MR scans were acquired at a unique institution with different scanners and clinical protocols, leading to different intensity distributions for a given MR sequence.

\textbf{3. RESECT-SEG:} RESECT-SEG is a small dataset of 23 3D brain \iUS with brain tumor delineations~\cite{resect-seg}. All images were acquired at a single institution with the same ultrasound probe. Note that this ultrasound probe differs from the one used in ReMIND, leading to a distribution gap between the \iUS in RESECT-SEG and ReMIND data.

\subsection{Implementation details}
\textbf{Network Architecture:} Given that raw brain ultrasound images are typically 2D, we adopted a 2D U-Net-based architecture. The spatial resolution and feature dimension of the coarsest latent variable ($z_{L}$) were chosen as $1\times1$ and $256$, respectively. The spatial and feature dimensions were successively doubled and halved after each level, resulting in a feature representation of dimension $8$ for each pixel at group $1$, denoted as $z_1\in\mathbb{R}^{192\times192\times8}$. This results in $L=7$ latent variable levels. Following state-of-the-art architectures~\cite{vahdat2020nvae}, residual cells from MobileNetV2~\cite{sandler2018mobilenetv2} are employed for the encoder and decoder, with Squeeze and Excitation~\cite{hu2018squeeze} and Swish activation. The image decoders $(\mu_{\theta_{j}^{x}})_{j=1}^{M}$ correspond to 5 ResNet blocks. At inference, we lower
the temperature of the parametric distributions to $0.5$, as performed in other HVAEs\cite{vahdat2020nvae}. The code is available at~\url{https://github.com/ReubenDo/MMHVAE}. 

\noindent \textbf{Training Parameters:} The models are trained for $1000$ epochs with a batch size of $B=16$. The parameters $\alpha_{\bm{r'}}^{(\bm{r})}$ are set such that the probability of drawing the full set of observable images (i.e., $\bm{r'}=\bm{r}$), the ultrasound image only, or a set of MRI is equal (i.e., $p=\frac{1}{3}$). Additionally, a uniform distribution is chosen over the set of subsets of observable MRI. At each training iteration, $T=3$ indicator vectors $\bm{r'}$ are drawn. The KL divergence is set to $\lambda_{KL}=0.001$. The weight of the GAN loss is set to $\lambda_{\text{GAN}}=0$ for the 800 first epochs and then to $\lambda_{\text{GAN}}=0.025$ for the last 200 epochs. Models were trained on an A100 40GB GPU.

\subsection{Competing methods}
We compared our approach to three state-of-the-art approaches for unified cross-modal image synthesis. Curriculum learning from our model was used to ensure standard sample selection for all competing methods. 

\noindent \textbf{MoPoE}~\cite{sutter2021generalized}: We implemented the Mixture-of-Products-of-Expert (MoPoE) using our network architecture for fair comparison. The unification strategy is the same as Ours with a single latent variable ($L=1$). 

\noindent \textbf{MM-GAN}~\cite{Sharma20}: MM-GAN is a unified synthesis model based on a convolutional GAN was considered. MM-GAN comprises CNN-based generator and discriminator networks, where the generator is based on U-Net. MM-GAN trains a single network under various source-target modality configurations. The original MM-GAN architecture was directly adopted. The unification strategy consists of concatenating all images and replacing missing ones with a zero tensor.

\noindent  \textbf{ResViT}~\cite{resvit}: The current state-of-the-art unified synthesis model using transformers. The original ResViT architecture was directly adopted. The unification strategy in ResViT uses zero-imputation as MM-GAN.

\noindent   \revmod{\textbf{M2DN}~\cite{meng2024multi}: The current state-of-the-art unified synthesis model based on diffusion models. M2DN applies a diffusion model with binary conditional codes to specify different imputation tasks. The unification strategy
consists of concatenating all images and replacing missing
ones with random noise. During the reverse diffusion step, the modalities are progressively reconstructed. To additionally perform harmonization, we adapted the original method using the Brownian bridge diffusion framework~\cite{li2023bbdm}.} 

\noindent   \revmod{\textbf{MHVAE}~\cite{dorent2023Unified}: MHVAE is our previously proposed model, which can handle only two modalities (iUS and \Ttwo). }

\begin{table*}[tb]
	\caption{\textbf{Comparison against the state-of-the-art unified models for image synthesis.} Mean and standard deviation values are presented. For a given input, values in bold indicate the methods that achieved the best performance and were found to be statistically significantly better than all others, based on a paired Wilcoxon signed-rank test with Bonferroni correction ($p < 0.01$). Arrows indicate the favorable direction of each metric.
	}\label{tab:Scores_synthesis}
	\resizebox{0.97\textwidth}{!}{
 \addtolength{\tabcolsep}{-0.5em}
	\begin{tabular}{l c *{18}{c}}
		\toprule
		\multirow{2}{*}{} & \multicolumn{4}{c}{\bf Input } & \multicolumn{3}{c}{\bf \iUS} & \multicolumn{3}{c}{\bf \Ttwo } & \multicolumn{3}{c}{\bf \Tone} & \multicolumn{3}{c}{\bf FLAIR}\\ 
       \cmidrule(lr){2-5} \cmidrule(lr){6-8} \cmidrule(lr){9-11} \cmidrule(lr){12-14} \cmidrule(lr){15-17}
       & \iUS & \text{T}\textsubscript{2} & \text{T}\textsubscript{1} & FL & PSNR(dB)$\uparrow$ & SSIM($\%$)$\uparrow$ & LPIPS($\%$)$\downarrow$  & PSNR(dB)$\uparrow$ & SSIM($\%$)$\uparrow$ & LPIPS($\%$)$\downarrow$ & PSNR(dB)$\uparrow$ & SSIM($\%$)$\uparrow$ & LPIPS($\%$)$\downarrow$ & PSNR(dB)$\uparrow$ & SSIM($\%$)$\uparrow$ & LPIPS($\%$)$\downarrow$  \\
		\midrule
\rowcolor{LightGray}
 \revmod{MoPoE~\cite{sutter2021generalized}} & $\bullet$ & $\circ$ & $\circ$ & $\circ$ &22.4 (3.6) & 75.7 (11.7) & 31.6 (13.1) &\textbf{21.8 (5.0)} & \textbf{83.0 (8.6)} & 24.1 (10.5) &\textbf{23.6 (5.3)} & \textbf{84.9 (7.6)} & 20.2 (9.5) &18.4 (8.2) & \textbf{79.3 (10.3)} & 22.9 (11.4) \\
\revmod{MHVAE~\cite{dorent2023Unified}} & $\bullet$ & $\circ$ & $\circ$ & $\circ$ &31.6 (2.2) & 92.4 (3.3) & 6.8 (2.6) &21.6 (4.0) & 80.7 (9.1) & 20.8 (8.8) & - &  - &  - & - &  - &  - \\
\rowcolor{LightGray}
MM-GAN \cite{Sharma20} & $\bullet$ & $\circ$ & $\circ$ & $\circ$ &25.3 (5.0) & 87.3 (7.3) & 10.8 (5.6) &19.5 (4.6) & 76.7 (11.0) & 19.9 (8.2) &21.3 (4.6) & 79.5 (9.4) & 17.0 (7.8) &16.7 (7.7) & 74.3 (11.9) & 19.3 (9.7) \\

ResViT \cite{resvit} & $\bullet$ & $\circ$ & $\circ$ & $\circ$ &32.5 (4.1) & 95.6 (2.6) & 3.2 (1.7) &20.7 (4.2) & 79.1 (9.8) & \textbf{16.0 (6.7)} &22.5 (5.0) & 81.7 (8.4) & \textbf{13.5 (6.6)} &17.8 (7.5) & 76.5 (10.8) & \textbf{14.5 (9.3)} \\

\rowcolor{LightGray}
\revmod{M2DN~\cite{meng2024multi}}  & $\bullet$ & $\circ$ & $\circ$ & $\circ$ &27.7 (5.5) & 88.0 (7.1) & 5.5 (2.8) &19.4 (5.1) & 80.1 (9.5) & 20.1 (8.9) &20.7 (5.7) & 83.0 (8.1) & 17.1 (8.3) &18.3 (7.5) & 78.6 (10.2) & 16.9 (10.0) \\
Ours & $\bullet$ & $\circ$ & $\circ$ & $\circ$ &\textbf{46.3 (2.9)} & \textbf{99.6 (0.5)} & \textbf{0.2 (0.4)} &20.8 (4.3) & 79.9 (9.6) & 16.3 (6.6) &22.2 (5.4) & 82.0 (8.5) & \textbf{14.0 (6.8)} &\textbf{18.6 (8.0)} & \textbf{79.0 (10.2)} & 17.2 (9.5) \\
\midrule
\rowcolor{LightGray}
 \revmod{MoPoE~\cite{sutter2021generalized}} & $\circ$ & $\bullet$ & $\circ$ & $\circ$ &\textbf{21.0 (4.4)} & \textbf{74.4 (12.4)} & 31.6 (13.7) &21.8 (5.1) & 83.2 (8.6) & 24.0 (10.5) &24.4 (4.5) & 85.9 (6.9) & 19.3 (9.2) &20.3 (4.8) & 81.0 (9.2) & 21.4 (9.9) \\
\revmod{MHVAE~\cite{dorent2023Unified}} & $\circ$ & $\bullet$ & $\circ$ & $\circ$ &20.8 (3.7) & 73.9 (12.0) & 20.5 (8.5) &28.6 (3.1) & 91.0 (3.9) & 5.7 (2.1) & - &  - &  - & - &  - &  - \\
\rowcolor{LightGray}
MM-GAN \cite{Sharma20} & $\circ$ & $\bullet$ & $\circ$ & $\circ$ &20.5 (4.4) & 73.3 (12.5) & 25.4 (10.6) &23.3 (4.4) & 93.0 (3.4) & 5.4 (2.1) &24.4 (3.5) & 86.1 (6.4) & 10.6 (4.9) &19.8 (3.8) & 79.3 (9.7) & 12.5 (5.4) \\
ResViT~\cite{resvit} & $\circ$ & $\bullet$ & $\circ$ & $\circ$ &20.0 (4.2) & 70.9 (13.3) & 19.3 (7.9) &28.6 (4.6) & 94.6 (3.0) & 5.0 (2.7) &\textbf{24.9 (4.1)} & \textbf{86.9 (6.4)} & \textbf{8.8 (4.5)} &22.4 (3.4) & 83.5 (7.7) & \textbf{8.5 (3.8)} \\
\rowcolor{LightGray}
 \revmod{M2DN~\cite{meng2024multi}}  & $\circ$ & $\bullet$ & $\circ$ & $\circ$ &20.6 (4.9) & \textbf{74.5 (12.4)} & 23.2 (9.3) &24.9 (5.3) & 95.0 (3.1) & 4.6 (2.5) &20.6 (4.8) & 84.9 (7.3) & 15.4 (7.5) &21.3 (4.4) & 82.1 (8.6) & 12.9 (6.1) \\
 
Ours & $\circ$ & $\bullet$ & $\circ$ & $\circ$ &20.6 (4.2) & 73.1 (12.4) & \textbf{18.9 (7.6)} &\textbf{35.9 (4.1)} & \textbf{98.5 (1.2)} & \textbf{1.1 (0.7)} &\textbf{24.9 (4.3)} & \textbf{86.7 (6.5)} & 9.0 (4.4) &\textbf{23.5 (4.1)} & \textbf{85.1 (7.3)} & 9.0 (4.4) \\
\rowcolor{LightGray}
\midrule
 \revmod{MoPoE~\cite{sutter2021generalized}} & $\circ$ & $\circ$ & $\bullet$ & $\circ$ &20.9 (4.5) & 73.8 (12.4) & 32.3 (13.6) &21.7 (5.0) & 82.8 (8.6) & 24.3 (10.5) &23.6 (5.1) & 85.0 (7.5) & 20.4 (9.5) &18.3 (8.2) & 79.2 (10.0) & 23.9 (11.0) \\
MM-GAN \cite{Sharma20} & $\circ$ & $\circ$ & $\bullet$ & $\circ$ &20.0 (4.2) & 72.4 (12.6) & 26.3 (10.6) &21.2 (4.5) & 82.1 (8.5) & 13.6 (5.5) &28.1 (4.0) & 94.8 (3.0) & 4.4 (2.2) &18.6 (6.7) & 78.1 (10.3) & 14.3 (7.6) \\
\rowcolor{LightGray}
ResViT~\cite{resvit} & $\circ$ & $\circ$ & $\bullet$ & $\circ$ &19.8 (4.2) & 70.0 (13.5) & \textbf{19.8 (8.0)} &22.6 (4.4) & 82.6 (8.2) & 12.1 (5.2) &29.6 (5.4) & 94.7 (3.1) & 4.3 (3.5) &20.1 (7.9) & 81.6 (8.5) & \textbf{10.6 (8.5)} \\
 \revmod{M2DN~\cite{meng2024multi}}  & $\circ$ & $\circ$ & $\bullet$ & $\circ$ &19.8 (5.0) & 73.1 (12.9) & 24.8 (9.7) &19.3 (5.3) & 80.4 (9.4) & 20.3 (9.0) &24.7 (7.5) & 93.5 (3.8) & 6.7 (3.9) &18.5 (8.0) & 78.9 (10.0) & 17.4 (10.2) \\
\rowcolor{LightGray}
Ours & $\circ$ & $\circ$ & $\bullet$ & $\circ$ &\textbf{21.1 (4.2)} & \textbf{74.0 (11.9)} & 20.9 (8.8) &\textbf{23.2 (4.4)} & \textbf{84.2 (7.9)} & \textbf{10.5 (4.6)} &\textbf{36.1 (6.2)} & \textbf{98.6 (2.0)} & \textbf{1.3 (3.0)} &\textbf{21.4 (7.8)} & \textbf{83.6 (7.9)} & \textbf{10.5 (7.3)} \\

\midrule
 \revmod{MoPoE~\cite{sutter2021generalized}} & $\circ$ & $\circ$ & $\circ$ & $\bullet$ &\textbf{20.9 (4.5)} & \textbf{74.8 (12.6)} & 31.2 (13.8) &22.0 (4.9) & \textbf{84.1 (8.0)} & 23.3 (10.1) &22.5 (7.2) & 84.9 (7.8) & 20.8 (10.2) &18.3 (7.8) & 79.5 (9.7) & 23.9 (11.5) \\
 
\rowcolor{LightGray}

MM-GAN \cite{Sharma20} & $\circ$ & $\circ$ & $\circ$ & $\bullet$ &20.5 (4.3) & 73.5 (12.8) & 25.4 (11.0) &20.5 (4.3) & 80.3 (9.4) & 15.9 (6.9) &22.4 (7.0) & 83.1 (8.4) & 13.4 (7.9) &22.2 (7.2) & 86.9 (7.3) & 9.3 (6.9) \\

ResViT~\cite{resvit} & $\circ$ & $\circ$ & $\circ$ & $\bullet$ &20.0 (4.3) & 71.3 (13.5) & \textbf{19.2 (8.0)} &\textbf{22.6 (4.2)} & 83.4 (8.0) & \textbf{11.5 (5.0)} &\textbf{22.7 (7.9)} & \textbf{85.2 (7.2)} & \textbf{10.6 (6.8)} &27.2 (8.6) & 95.0 (5.2) & 4.2 (8.3) \\

\rowcolor{LightGray}
 \revmod{M2DN~\cite{meng2024multi}}  & $\circ$ & $\circ$ & $\circ$ & $\bullet$ &20.6 (5.0) & \textbf{74.8 (12.7)} & 22.8 (9.3) &18.6 (5.2) & 81.2 (9.0) & 18.4 (8.1) &18.5 (8.2) & 83.3 (8.2) & 16.8 (8.4) &22.0 (8.7) & 94.5 (4.7) & 4.5 (6.3) \\
Ours & $\circ$ & $\circ$ & $\circ$ & $\bullet$ &20.3 (4.4) & 73.4 (12.9) & 20.7 (9.1) &21.7 (4.4) & 83.6 (8.1) & 11.9 (4.8) &22.4 (7.5) & 84.1 (7.9) & 10.9 (7.4) &\textbf{31.3 (8.6)} & \textbf{97.6 (5.1)} & \textbf{2.5 (7.6)} \\
\midrule

\rowcolor{LightGray}
 \revmod{MoPoE~\cite{sutter2021generalized}} & $\circ$ & $\bullet$ & $\bullet$ & $\circ$ &\textbf{21.0 (4.5)} & \textbf{74.3 (12.4)} & 31.9 (13.6) &21.8 (5.0) & 82.9 (8.6) & 24.3 (10.5) &24.4 (4.4) & 85.9 (6.9) & 19.3 (9.2) &20.2 (4.9) & 80.3 (9.3) & 22.1 (9.9) \\
MM-GAN \cite{Sharma20} & $\circ$ & $\bullet$ & $\bullet$ & $\circ$ &20.7 (4.4) & 73.6 (12.4) & 24.9 (10.4) &24.0 (4.4) & 93.3 (3.3) & 5.1 (2.1) &28.2 (3.1) & 93.9 (2.7) & 5.0 (2.0) &20.4 (3.8) & 80.6 (9.3) & 12.0 (5.3) \\
\rowcolor{LightGray}
ResViT~\cite{resvit} & $\circ$ & $\bullet$ & $\bullet$ & $\circ$ &20.1 (4.3) & 71.1 (13.2) & 19.0 (7.8) &28.2 (4.7) & 93.9 (3.3) & 5.2 (2.9) &29.2 (3.9) & 93.6 (3.2) & 4.6 (2.6) &22.8 (3.4) & 84.6 (7.2) & \textbf{7.6 (3.5)} \\
 \revmod{M2DN~\cite{meng2024multi}}  & $\circ$ & $\bullet$ & $\bullet$ & $\circ$ &20.6 (4.8) & \textbf{74.4 (12.2)} & 23.0 (9.2) &24.9 (5.4) & 95.0 (3.0) & 4.5 (2.3) &24.9 (7.2) & 93.9 (3.7) & 5.4 (2.7) &21.6 (4.3) & 82.7 (8.4) & 12.1 (5.7) \\
\rowcolor{LightGray}
Ours & $\circ$ & $\bullet$ & $\bullet$ & $\circ$ &\textbf{20.9 (4.0)} & 73.6 (12.0) & \textbf{18.4 (7.4)} &\textbf{35.6 (4.5)} & \textbf{98.4 (1.3)} & \textbf{1.1 (0.7)} &\textbf{36.1 (4.6)} & \textbf{98.7 (0.7)} & \textbf{1.1 (0.6)} &\textbf{23.9 (3.8)} & \textbf{85.8 (6.9)} & 8.6 (4.2) \\

\midrule
 \revmod{MoPoE~\cite{sutter2021generalized}} & $\circ$ & $\bullet$ & $\circ$ & $\bullet$ &\textbf{21.3 (4.6)} & 75.5 (12.5) & 30.9 (14.0) &22.3 (4.9) & 84.3 (8.0) & 23.3 (10.1) &24.2 (4.4) & 86.3 (6.6) & 18.5 (8.5) &20.2 (4.7) & 80.9 (9.2) & 21.4 (9.8) \\
\rowcolor{LightGray}
MM-GAN \cite{Sharma20} & $\circ$ & $\bullet$ & $\circ$ & $\bullet$ &21.1 (4.5) & 74.9 (12.3) & 24.4 (10.9) &24.1 (4.2) & 93.3 (3.3) & 4.7 (2.0) &24.9 (3.5) & 87.1 (6.1) & 9.3 (4.3) &25.7 (4.2) & 91.2 (4.6) & 5.5 (2.5) \\
ResViT~\cite{resvit} & $\circ$ & $\bullet$ & $\circ$ & $\bullet$ &20.3 (4.3) & 72.4 (13.2) & 18.5 (7.9) &28.3 (4.5) & 94.0 (3.2) & 5.3 (3.0) &\textbf{25.1 (3.9)} & \textbf{88.2 (5.9)} & \textbf{7.8 (3.8)} &28.3 (3.6) & 95.2 (2.4) & 3.2 (2.1) \\
\rowcolor{LightGray}
 \revmod{M2DN~\cite{meng2024multi}}  & $\circ$ & $\bullet$ & $\circ$ & $\bullet$ &\textbf{21.2 (5.0)} & \textbf{75.8 (12.4)} & 21.7 (9.1) &25.5 (5.1) & 95.5 (2.9) & 3.8 (2.0) &19.8 (5.3) & 85.8 (7.0) & 13.6 (6.1) &22.6 (6.9) & 95.2 (2.6) & 3.4 (1.9) \\
Ours & $\circ$ & $\bullet$ & $\circ$ & $\bullet$ &20.4 (4.0) & 73.7 (12.5) & \textbf{18.1 (7.6)} &\textbf{35.1 (4.2)} & \textbf{98.3 (1.5)} &\textbf{ 1.4 (0.9)} &24.6 (3.9) & 87.0 (6.1) & 8.8 (4.0) &\textbf{33.3 (4.0)} & \textbf{98.2 (1.1)} & \textbf{1.3 (0.8)} \\

\midrule
\rowcolor{LightGray}
 \revmod{MoPoE~\cite{sutter2021generalized}} & $\circ$ & $\circ$ & $\bullet$ & $\bullet$ &\textbf{20.9 (4.6)} & \textbf{74.4 (12.6)} & 31.8 (13.9) &22.0 (4.7) & 83.6 (8.1) & 24.1 (10.1) &22.8 (7.0) & 85.2 (7.8) & 20.3 (8.8) &18.2 (7.9) & 79.1 (9.7) & 24.3 (10.4) \\
MM-GAN \cite{Sharma20} & $\circ$ & $\circ$ & $\bullet$ & $\bullet$ &20.5 (4.2) & 73.4 (12.6) & 25.3 (10.7) &21.4 (4.3) & 82.2 (8.2) & 14.9 (6.3) &26.9 (6.8) & 93.2 (5.8) & 5.3 (5.8) &23.0 (7.5) & 89.0 (6.7) & 7.7 (6.9) \\
\rowcolor{LightGray}
ResViT~\cite{resvit} & $\circ$ & $\circ$ & $\bullet$ & $\bullet$ &20.0 (4.3) & 71.2 (13.3) & \textbf{19.1 (7.9)} &23.4 (4.4) & 84.7 (7.3) & 10.4 (4.7) &27.2 (8.6) & 93.1 (4.5) & 5.6 (6.3) &26.3 (9.0) & 94.2 (5.5) & 4.7 (8.8) \\
 \revmod{M2DN~\cite{meng2024multi}}  & $\circ$ & $\circ$ & $\bullet$ & $\bullet$ &20.6 (4.9) & \textbf{74.4 (12.5)} & 22.9 (9.2) &18.5 (5.2) & 81.7 (8.7) & 17.1 (7.4) &22.7 (9.4) & 93.7 (4.8) & 5.7 (6.0) &22.4 (9.0) & 94.6 (4.9) & 4.5 (6.8) \\
\rowcolor{LightGray}
Ours & $\circ$ & $\circ$ & $\bullet$ & $\bullet$ &\textbf{20.9 (4.0)} & 74.0 (12.0) & 19.4 (8.3) &\textbf{23.9 (4.0)} & \textbf{85.6 (6.9)} & \textbf{9.5 (3.8)} & \textbf{33.5 (9.8)} & \textbf{97.9 (4.0)} & \textbf{2.1 (6.2)} &\textbf{31.1 (8.9)} & \textbf{97.6 (4.0)} & \textbf{2.4 (6.7) }\\
\midrule
 \revmod{MoPoE~\cite{sutter2021generalized}} & $\circ$ & $\bullet$ & $\bullet$ & $\bullet$ &\textbf{21.2 (4.6)} & \textbf{75.0 (12.4)} & 31.5 (13.9) &22.1 (4.9) & 83.6 (8.1) & 24.0 (10.1) &24.3 (4.3) & 86.4 (6.6) & 18.5 (8.4) &20.2 (4.7) & 80.3 (9.2) & 22.0 (9.8) \\
\rowcolor{LightGray}
MM-GAN \cite{Sharma20} & $\circ$ & $\bullet$ & $\bullet$ & $\bullet$ &\textbf{21.1 (4.4)} & 74.6 (12.2) & 24.4 (10.7) &25.3 (4.0) & 93.8 (3.0) & 4.5 (1.9) &28.1 (3.5) & 93.6 (3.0) & 4.8 (2.0) &26.2 (3.9) & 92.2 (4.1) & 5.2 (2.4) \\
ResViT~\cite{resvit} & $\circ$ & $\bullet$ & $\bullet$ & $\bullet$ &20.3 (4.3) & 72.1 (13.0) & 18.6 (7.8) &27.9 (4.7) & 93.5 (3.4) & 5.6 (3.2) &28.2 (3.9) & 93.0 (3.5) & 4.8 (2.4) &27.8 (3.5) & 94.7 (2.7) & 3.5 (2.2) \\
\rowcolor{LightGray}
 \revmod{M2DN~\cite{meng2024multi}}  & $\circ$ & $\bullet$ & $\bullet$ & $\bullet$ &21.0 (4.9) & \textbf{75.1 (12.3)} & 22.0 (9.0) &25.9 (5.0) & 95.5 (2.5) & 3.9 (1.9) &23.6 (6.9) & 94.3 (3.5) & 4.7 (2.4) &23.2 (6.7) & 95.4 (2.3) & 3.4 (1.8) \\
Ours & $\circ$ & $\bullet$ & $\bullet$ & $\bullet$ &20.6 (3.9) & 73.9 (12.1) & \textbf{18.0 (7.4)} &\textbf{35.1 (4.3)} &\textbf{ 98.2 (1.6)} & \textbf{1.3 (0.8)} &\textbf{34.2 (5.0)} & \textbf{98.4 (1.0)} & \textbf{1.2 (0.7)} &\textbf{32.8 (4.3)} & \textbf{98.1 (1.1)} & \textbf{1.5 (0.8)} \\
\midrule
\rowcolor{LightGray}
  & \multicolumn{4}{c}{ \revmod{MoPoE~\cite{sutter2021generalized}}} &21.3 (4.3) & 74.6 (12.3) & 31.7 (13.5) &21.8 (5.0) & 83.2 (8.5) & 24.1 (10.4) &23.8 (5.2) & 85.4 (7.3) & 19.8 (9.3) &19.2 (6.8) & 79.9 (9.6) & 22.8 (10.6) \\

  & \multicolumn{4}{c}{MM-GAN \cite{Sharma20}} &21.6 (4.9) & 76.7 (13.0) & 21.9 (11.5) &22.2 (4.8) & 86.5 (10.0) & 10.8 (7.9) &25.4 (5.2) & 88.5 (8.8) & 9.3 (7.2) &21.4 (6.8) & 83.7 (10.6) & 11.0 (7.9) \\
  \rowcolor{LightGray}
  & \multicolumn{4}{c}{ResViT~\cite{resvit}} &23.0 (6.8) & 76.9 (15.8) & 15.4 (9.8) &25.1 (5.5) & 87.8 (9.2) & 9.3 (6.3) &26.3 (5.9) & 89.1 (7.8) & 7.9 (6.1) &24.0 (7.5) & 88.0 (10.0) & 7.2 (7.8) \\

  & \multicolumn{4}{c}{ \revmod{M2DN~\cite{meng2024multi}} } &22.2 (6.0) & 77.5 (12.9) & 19.1 (11.3) &22.1 (6.0) & 87.8 (10.1) & 12.0 (10.0) &22.4 (7.0) & 88.8 (7.9) & 11.1 (8.1) &21.1 (7.5) & 87.7 (10.3) & 9.5 (9.0) \\
  \rowcolor{LightGray}
\multirow{-5}{*}{\rotatebox[origin=c]{90}{\bf Average}} & \multicolumn{4}{c}{Ours} &\textbf{27.0 (11.6)} & \textbf{79.9 (15.4)} & \textbf{14.8 (10.9)} &\textbf{28.9 (8.0)} & \textbf{90.5 (10.1)} & \textbf{7.0 (7.3)} &\textbf{29.5 (8.5)} & \textbf{91.4 (9.2)} & \textbf{6.4 (7.2)} &\textbf{26.9 (8.7)} & \textbf{90.5 (9.9)} & \textbf{6.8 (8.3)} \\
\arrayrulecolor{black}\bottomrule
	\end{tabular}
	}
\end{table*}

\subsection{Experiments design}
To evaluate the effectiveness of our framework and compare it with state-of-the-art unified approaches, we conducted experiments on three tasks: (i) cross-modal image synthesis using paired data, (ii) downstream tumor segmentation, and (iii) downstream image registration. We also performed several ablation studies.

\subsubsection{Harmonized cross-modal image synthesis}This task aims to synthesize harmonized images from all $M=4$ modalities given any incomplete set of paired non-harmonized images. Specifically, if the target image is available as input, the task is to perform image harmonization. If the target image is missing, the task is to synthesize the harmonized version of the missing image.

Medical data often presents significant intensity distribution shifts when acquired at different centers with varying imaging protocols (e.g., different scanners and acquisition parameters). For example, these intra-modality shifts can be observed for the MR data from ReMIND, as shown in Appendix Fig. 1.  Through preliminary experiments, we found that performing cross-modality synthesis without considering these shifts leads to the synthesis of blurry MR images, mainly due to the intrinsic one-to-many contrast mapping between input and target images. To address this issue, data harmonization is required.

For data harmonization, we applied contrast linear normalization. We first shifted the minimal intensity value to $0$. Then, we aligned the median intensity values within white matter, estimated using SynthSeg~\cite{billot_synthseg_2023}, to  $\frac{1}{7}$, $\frac{1}{5}$ and $\frac{1}{3}$ for \Ttwo, \Tone and FLAIR respectively,  As shown in Figure~\ref{fig:distribution}, this method aligns well the intensity distribution when median values are obtained from the 3D volumes. However, significant shifts were observed when normalization was performed slice by slice. Since our model takes 2D slices as input, we applied 3D contrast normalization only to the ground truth target images during training and testing, but not to the 2D input slices. Therefore, when a 2D MR slice is available as input, the task is to harmonize it.

Ultrasound images, on the other hand, depend on the probe's position and angle, as they are generated from the reflection, scatter, and absorption of sound waves through tissues. To account for this spatial dependency during synthesis, all MR images were cropped to the field of view of the \iUS probe, allowing us to encode the probe’s position and angle relative to the brain surface.

Since paired data is available for evaluation (ReMIND), we employed standard supervised evaluation metrics: PSNR (Peak Signal-to-Noise Ratio), SSIM (Structural Similarity), and LPIPS~\cite{zhang2018perceptual} (Learned Perceptual Image Patch Similarity). We conducted a 3-fold cross-validation using stratified sampling to ensure that all possible combinations of inputs are adequately represented in each test set.

\subsubsection{Ablation study}
We performed an ablation study to assess the impact of three key components: (i) the hierarchical latent representation, (ii) the probabilistic fusion strategy for missing images, and (iii) the GAN-based regularization. These experiments were conducted on one fold.


First, we assessed the utility of the hierarchy in the latent representation. We compared our approach with $L=7$ latent variables to a non-hierarchical approach $L=1$ (group 7). Moreover, we compared our approach to three hierarchical approaches with $L=2$ (groups 1 and 7), $L=3$ (groups 1, 4, and 7), and $L=5$ (groups 1, 3, 4, 5, and 7) latent variables. A quantitative assessment was performed for the cross-modal image synthesis task presented above.

Second, we compared our proposed fusion operation with two common techniques for handling missing data. 

\noindent \textbf{Concat. w/ zeros} uses the concatenation operation to merge information from multiple images. Similarly to MM-GAN~\cite{Sharma20} and ResViT~\cite{resvit}, zero imputation is used when images are missing. In this model, only one encoder is used. 

\noindent \textbf{Average} is a deterministic variant of our framework using the same network architecture. Features are extracted independently from each available image and averaged at each level of the hierarchy, as in~\cite{hemis,pimms,dorent2021learning}. 

Third, we evaluated the relevance of our GAN-based regularization by training our framework with and without ($\lambda_{GAN}=0$) it on the cross-modal image synthesis. Quantitative and qualitative assessments were performed.

\begin{figure*}[tb!]
    \centering
    \includegraphics[width=0.93\textwidth]{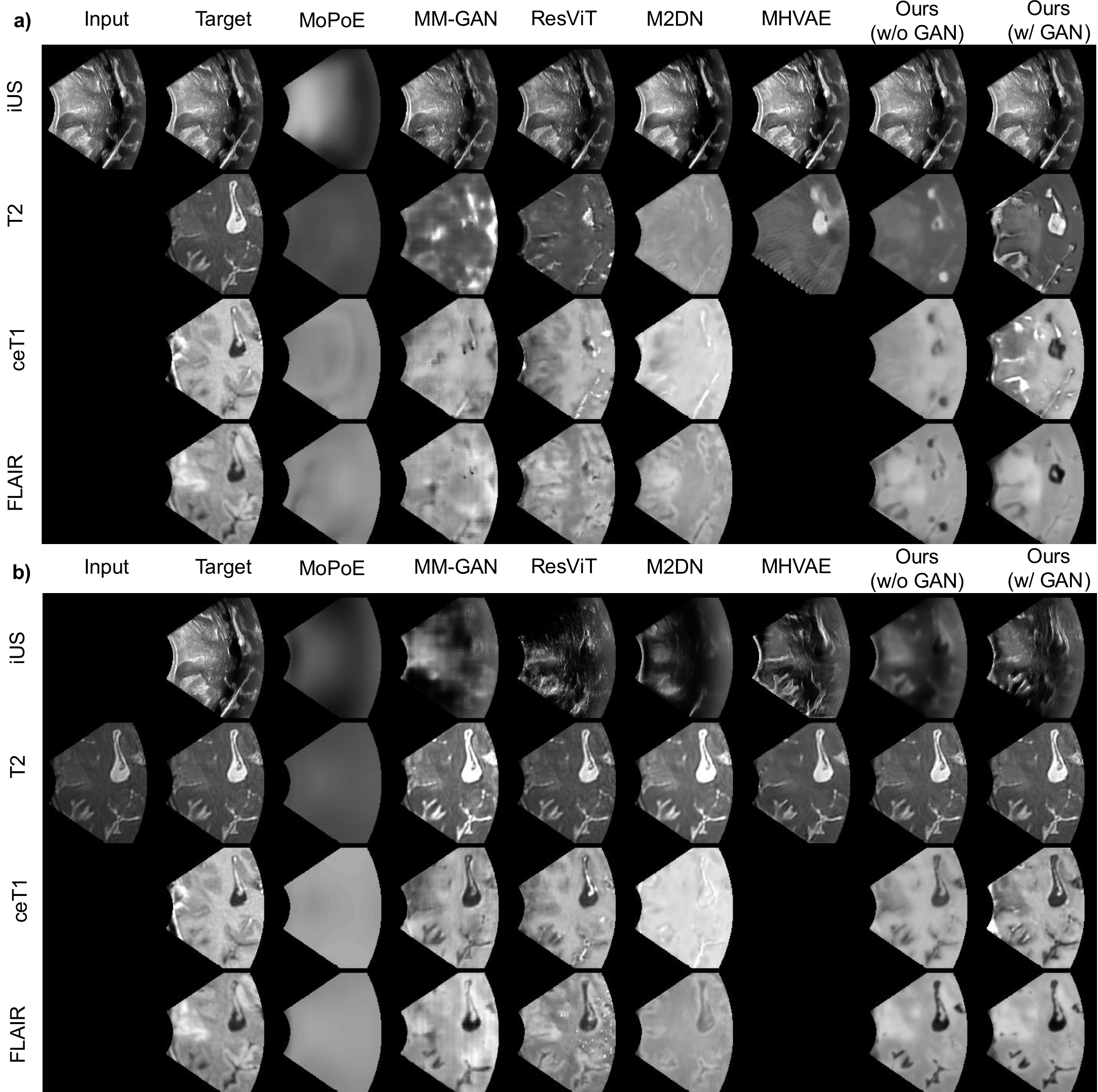}
    \caption{Qualitative comparison of our method with all competing methods for synthesizing all modalities (iUS, \Ttwo, \Tone, FLAIR) from (a) iUS; (b) \Ttwo. Our approach generates sharper images with better contrast differentiation between tissues and modality-specific patterns (e.g. speckles for iUS).}
    \label{fig:qualitative_results_synthesis}
\end{figure*}

\subsubsection{Brain tumor segmentation in \iUS} To further compare our proposed framework with competing methods, we conducted another set of experiments on the challenging downstream task of brain tumor segmentation in ultrasound images. While automatic segmentation of brain tumors in \iUS could provide intraoperative guidance during surgery, annotating ultrasound images requires time and rare clinical expertise as tumor boundaries are often unclear and ambiguous. For this reason, only a very small annotated \iUS dataset RESECT-SEG (N=22 annotated 3D iUS) is currently available. In contrast, the performance of deep learning models for brain tumor segmentation in MR images has reached comparable performance to human annotators~\cite{bakas2019identifying} thanks to the availability of large datasets, such as UPenn-GBM (N=611). In this task, we propose to leverage our cross-modal synthesis framework to learn to perform brain tumor segmentation in \iUS by synthesizing \iUS from MR data and exploiting available MR annotations, i.e., performing
cross-modality domain adaptation for image segmentation~\cite{dorent2023crossmoda}.

First, we generate virtual 3D ultrasound sweeps in pre-operative MR data from UPenn-GBM using~\cite{dorent2024patientspecific}. Second, we exploit MM-GAN, ResViT,  \revmod{M2DN}, and Ours from the synthesis task to synthesize \iUS data from 2D MR slices in the virtual sweeps’ field of view. This created a paired dataset of synthetic ultrasound images and brain tumor annotations for each framework. Third, we trained a 5-fold nnU-Net ensemble~\cite{isensee2021nnu}, a well-established framework for 3D image segmentation, to perform automated segmentation in the ultrasound images using each synthetic paired dataset. This led to the creation of one ensemble for each cross-modal synthesis technique.   \revmod{To further benchmark performance against a fully supervised baseline, we also trained a 5-fold nnU-Net ensemble on the only publicly available dataset (RESECT-SEG) containing real iUS with tumor annotations.} Finally, we assessed the performance of the trained ensembles using the RESECT-SEG dataset with $N=22$ annotated ultrasound images and on a subset of N=6 ReMIND \iUS images not used during training~\cite{dorent2024223}. Dice Score Coefficient (DSC) and Average Symmetric Surface Distance (ASSD) were computed to compare manual annotations with predictions.

\subsubsection{MR-\iUS image registration} 
Finally, we conducted experiments on the problem of multimodal image registration between pre-operative MR and 3D \iUS images. Registering MRI and \iUS is challenging because they provide different characterizations of tissues associated with different physical properties. Consequently, MRI and \iUS provide complementary information in very different contrasts. To facilitate the registration process, we investigate using our cross-modal synthesis framework. Specifically, we propose registering each pre-operative MRI with its corresponding synthesized sequence from \iUS. 
As we have access to paired data, we simulated rigid deformations within four ranges with median displacements of 0-4 mm, 4-8 mm, 8-12 mm, and 12-16 mm. Then, registration was performed using either the original \iUS volume or the simulated MR sequence in the field of view of the \iUS using MM-GAN, ResViT, \revmod{M2DN}, and Ours. Gradient-descent registration was used to optimize the 6 degrees of freedom using a multiscale normalized local cross-correlation metric~\cite{lapirn}. We reported the average Target Registration Error (TRE) in mm for each MR sequence 
\revmod{by comparing true and predicted displacement at each voxel.}

\section{Experimental results}
This section presents the results of the harmonized cross-modal synthesis task, ablation study and downstream tasks.

\begin{table*}[tb]
\caption{\textbf{Ablation studies:} a) effect of number of latent variables, b) fusing operation, and c) GAN loss. Scores are averaged across all possible combinations of inputs. Values in bold indicate the best performance based on a paired Wilcoxon signed-rank test with Bonferroni correction ($p<0.01$). Experiments performed on one fold of the ReMIND dataset.}
\label{tab:Scores_ablation_all}
\resizebox{0.99\textwidth}{!}{
\addtolength{\tabcolsep}{-0.5em}
\begin{tabular}{l c *{12}{c}}
\toprule
\multicolumn{1}{c}{\bf Setting} & \multicolumn{3}{c}{\bf \iUS} & \multicolumn{3}{c}{\bf \Ttwo} & \multicolumn{3}{c}{\bf \Tone} & \multicolumn{3}{c}{\bf FLAIR} \\
\cmidrule(lr){1-1} \cmidrule(lr){2-4} \cmidrule(lr){5-7} \cmidrule(lr){8-10} \cmidrule(lr){11-13}
& PSNR(dB)$\uparrow$ & SSIM(\%)$\uparrow$ & LPIPS(\%)$\downarrow$ & PSNR(dB)$\uparrow$ & SSIM(\%)$\uparrow$ & LPIPS(\%)$\downarrow$ & PSNR(dB)$\uparrow$ & SSIM(\%)$\uparrow$ & LPIPS(\%)$\downarrow$ & PSNR(dB)$\uparrow$ & SSIM(\%)$\uparrow$ & LPIPS(\%)$\downarrow$ \\
\midrule
\multicolumn{13}{l}{\textbf{a) Number of latent variables $L$}} \\
\rowcolor{LightGray}
 $L=1$ &21.5 (4.5) & 75.6 (12.3) & 30.7 (13.8) &21.6 (4.9) & 83.4 (8.3) & 24.1 (10.5) &24.0 (4.4) & 86.0 (6.9) & 19.2 (9.1) &19.2 (4.6) & 80.4 (9.9) & 22.9 (10.7) \\
 $L=2$ &21.9 (4.3) & 75.5 (11.9) & 21.4 (10.2) &22.6 (4.5) & 81.4 (9.4) & 15.2 (6.7) &24.6 (4.1) & 84.0 (7.9) & 11.8 (5.8) &21.1 (4.5) & 80.9 (9.7) & 16.4 (7.8) \\
\rowcolor{LightGray}
 $L=3$ &22.8 (4.5) & 77.3 (11.2) & 19.3 (10.3) &24.6 (4.8) & 85.0 (8.3) & 11.3 (5.8) &25.7 (4.2) & 87.0 (7.4) & 9.7 (5.0) &23.7 (4.7) & 85.1 (8.1) & 10.9 (6.0) \\
 $L=5$ &23.5 (6.1) & 78.4 (13.4) & 15.0 (8.2) &27.8 (6.9) & 89.8 (9.3) & 8.0 (6.4) &28.5 (6.4) & 90.6 (8.5) & 6.5 (5.9) &26.8 (6.6) & 89.4 (9.6) & \textbf{6.5 (5.7)} \\
\rowcolor{LightGray}
 $L=7$ &\textbf{27.0 (11.5)} & \textbf{80.6 (15.0)} & \textbf{14.7 (10.9)} &\textbf{28.6 (7.8)} & \textbf{90.7 (9.7)} & \textbf{7.1 (7.1)} &\textbf{29.8 (7.6)} & \textbf{91.7 (8.9)} & \textbf{6.2 (6.3)} &\textbf{27.4 (7.0)} & \textbf{91.0 (9.3)} & 6.7 (7.2) \\
\midrule
\multicolumn{13}{l}{\textbf{b) Fusing operation}} \\
\rowcolor{LightGray}
 Concat. w/ zeros &26.1 (9.6) & \textbf{80.7 (14.8)} & 15.6 (11.2) &25.8 (5.9) & 88.6 (10.4) & 9.0 (7.1) &26.3 (5.2) & 90.1 (8.8) & 8.0 (6.0) &24.0 (4.8) & 88.2 (9.6) & 8.7 (7.0) \\
Average  & 26.3 (10.1) & 80.2 (15.2) & \textbf{14.4 (10.2)} &27.0 (7.2) & 89.0 (10.6) & 8.7 (7.8) &27.2 (6.0) & 90.0 (9.1) & 7.6 (6.4) &25.7 (6.2) & 89.7 (9.5) & 7.5 (7.3) \\
\rowcolor{LightGray}
Ours &\textbf{27.0 (11.5)} & \textbf{80.6 (15.0)} & \textbf{14.7 (10.9)} &\textbf{28.6 (7.8)} & \textbf{90.7 (9.7)} & \textbf{7.1 (7.1)} &\textbf{29.8 (7.6)} &\textbf{ 91.7 (8.9)} & \textbf{6.2 (6.3)} &\textbf{27.4 (7.0)} &\textbf{ 91.0 (9.3)} & \textbf{6.7 (7.2)} \\
\midrule
\multicolumn{13}{l}{\textbf{c) GAN loss}} \\
\rowcolor{LightGray}
Without GAN  & \textbf{28.1 (12.0)} & \textbf{81.4 (14.5}) & 20.6 (15.3) &\textbf{30.1 (8.0)} & \textbf{91.6 (9.0)} & 8.3 (9.5) &3\textbf{0.6 (8.3)} & \textbf{92.6 (7.9)} & 7.4 (8.8) &\textbf{27.2 (8.8)} & \textbf{90.5 (9.8)} & \textbf{6.6 (8.3)} \\
With GAN &27.0 (11.5) & 79.7 (15.6) & \textbf{14.6 (10.9)} &28.8 (8.0) & 89.8 (10.8) & \textbf{7.1 (7.4)} &29.4 (8.4) & 90.6 (9.9) & \textbf{6.3 (7.0)} &26.8 (8.6) & 90.1 (10.2) & \textbf{6.6 (8.1)} \\
\bottomrule
\end{tabular}
}
\end{table*}

\subsection{Harmonized cross-modal image synthesis results}
We first assessed the effectiveness of MMHVAE in performing harmonized cross-modal image synthesis of multi-sequence MRI and ultrasound. MMHVAE was compared against the state-of-the-art unified synthesis framework using convolutional (MoPoE, MM-GAN), transformer (ResVIT),  \revmod{and diffusion models (M2DN).} Performance on unified models was evaluated at test time for each possible combination of input images in the ReMIND dataset. Qualitative results are shown in Figure~\ref{fig:qualitative_results_synthesis}, and detailed quantitative results are presented in Table~\ref{tab:Scores_synthesis}.

\textbf{Harmonization task:} When the target MR modality is present in the input set (indicated by a bullet point), the task is to perform image harmonization. Our method consistently outperforms the competing models across all target MR sequence modalities (\Ttwo, \Tone, and FLAIR). For instance, when \Ttwo harmonization is performed (non-harmonized \Ttwo present in input), our method achieves an SSIM greater than $98.2\%$ and a lower LPIPS than $1.3\%$ for any combination, indicating near-perfect structural similarity and perceptual quality. In contrast, MoPoE, MM-GAN, ResViT and  \revmod{M2DN} achieve lower SSIM scores ($82.9\%$, $93.0\%$, $93.5\%$, and \revmod{$95.0\%$} respectively) and higher LPIPS values ($24.3\%$, $5.4\%$, $5.6\%$, and \revmod{$4.6\%$} respectively). Similar conclusions can be drawn for ceT1 and FLAIR modalities, where our approach achieves the highest PSNR, SSIM, and the lowest LPIPS.  \revmod{Finally, our approach also significantly outperforms our previously proposed approach, MHVAE, on \Ttwo harmonization ($98.5\%$ vs $91.0\%$ SSIM; $5.7$ vs $1.1$ LPIPS).} This set of experiences demonstrates the superiority of our method in harmonizing images with minimal loss of information.

\textbf{Cross-modal synthesis task:} When the target modality is missing in the input set (indicated by an empty circle), the task synthesizes the harmonized missing modality using the available input images. Figure~\ref{fig:qualitative_results_synthesis} shows qualitative results for various input combinations. Our method consistently outperforms state-of-the-art approaches across all modalities. First, as illustrated in Figure~\ref{fig:qualitative_results_synthesis}, MoPoE produces noticeably blurry images, highlighting its limitations in generating high-quality outputs. In contrast, our approach produces more realistic and detailed images than other unified synthesis methods. Specifically, our method generates synthetic ultrasound images with more realistic textures, such as the presence of speckles, while also better preserving anatomical structures. These qualitative improvements align with the quantitative metrics shown in Table~\ref{tab:Scores_synthesis}, where our method achieves lower perceptual LPIPS and higher structural SSIM scores.

Although MM-GAN, ResViT, and  \revmod{M2DN} were primarily designed for synthesizing multi-parametric MR scans, our approach proves more effective at performing cross-modal synthesis between different MR sequences. For instance, our method reconstructs FLAIR scans more accurately from \Tone, \Ttwo, or a combination of both, compared to these two methods. This shows that while our model learns a shared latent representation, it outperforms state-of-the-art methods that rely on advanced network architectures .

\revmod{
\textbf{Computational resources: } Finally, our approach is significantly more computationally efficient than the best-performing baseline, ResViT, with a much lower time complexity (13G vs. 274G MACs) and a smaller model size (14M vs. 293M parameters). Compared to M2DN, which requires 18 inference steps, our method is both faster (55 ms vs. 290 ms) and more efficient (13G vs. 24G MACs). Moreover, our hierarchical fusion strategy introduces only a marginal overhead compared to MoPoE, with a $19\%$ increase in time complexity and a $4\%$ increase in model size.}

\subsection{Ablation study results}
\subsubsection{Number of levels in the hierarchy}

The results of the ablation study, presented in Table~\ref{tab:Scores_ablation_all}, clearly demonstrate the positive impact of increasing the number of latent variables $L$ in the hierarchical representation on the performance of our cross-modal image synthesis framework. As $L$ increases from 1 to 7, there is a consistent improvement in the quality of the synthesized images across all modalities (\iUS, \Ttwo, \Tone, and FLAIR). For instance, with $L=2$, there is already an improvement over the non-hierarchical approach ($L=1$), with an increase in PSNR values across modalities, particularly for \Ttwo (22.6 dB for $L=2$ vs. 21.6 dB for $L=1$) and FLAIR (21.1 dB for $L=2$ vs. 19.2 dB for $L=1$). As the number of latent variables increases, these improvements become evident. For example, with $L=5$, the PSNR for \Ttwo and FLAIR increases to 27.8 dB and 26.8 dB.

The model with $L=7$ latent variables achieves the highest performance across all metrics, with a PSNR of 27.0 dB for \iUS, 28.6 dB for \Ttwo, 29.8 dB for \Tone, and 27.4 dB for FLAIR. Similarly, the SSIM scores are also highest and the LPIPS values are the lowest, indicating better structural similarity and perceptual quality.  In short, as the latent structure becomes more complex with additional variables, the model gains in expressiveness, leading to more accurate and realistic image synthesis.

\begin{table*}[tb]
	\caption{\textbf{Comparison against the state-of-the-art unified models on the image registration downstream task } between \iUS and rigidly deformed (a) \Ttwo, (b) \Tone and (c) FLAIR scans. The registration algorithm either uses the acquired \iUS as a reference image or the translated \iUS in the modality of the moving image. Best performance in bold is based on a paired Wilcoxon signed-rank test ($p<0.05$).  Mean and standard deviation of Target Registration Error are given in mm for four ranges of displacements.
	}\label{tab:Scores_reg_t1}
 	\resizebox{0.33\textwidth}{!}{
 \addtolength{\tabcolsep}{-0.5em}
	\begin{tabular}{l l *{6}{c}}
    \multicolumn{6}{c}{\bf (a) iUS to \Ttwo registration } \\
		\toprule
  \rowcolor{white}
		&  & \multicolumn{1}{c}{\bf 0-4 mm } & \multicolumn{1}{c}{\bf 4-8 mm} & \multicolumn{1}{c}{\bf 8-12 mm } & \multicolumn{1}{c}{\bf 12-16 mm} \\ 
		\midrule
\rowcolor{LightGray}
\midrule
\multicolumn{2}{l}{Acquired iUS}  & 1.0 (0.9) & \textbf{1.5 (1.7)} & \textbf{2.0 (2.7)} & \textbf{7.3 (7.3)} \\
\midrule
 &MM-GAN~\cite{Sharma20} & 1.2 (0.8) & 2.0 (2.5) & 4.4 (5.3) & 9.6 (7.3) \\
 \rowcolor{LightGray}
 &ResViT~\cite{resvit}  & \textbf{0.8 (0.3)} & \textbf{1.4 (2.4)} & \textbf{2.5 (4.0)} & \textbf{7.1 (7.2)} \\
 & \revmod{M2DN~\cite{meng2024multi}}  & 1.1 (0.7) & 3.5 (4.4) & 5.8 (5.3) & \textbf{9.2 (6.1)} \\
 \rowcolor{LightGray}
\multirow{-4}{*}{\rotatebox[origin=c]{90}{\textbf{Tran. \Ttwo}} } &
 Ours & \textbf{0.7 (0.2)} & \textbf{1.3 (2.3)} & \textbf{2.0 (3.4)} & \textbf{7.3 (6.9)}\\
\arrayrulecolor{black}\bottomrule
	\end{tabular}
 }
 	\resizebox{0.34\textwidth}{!}{
 \addtolength{\tabcolsep}{-0.5em}
	\begin{tabular}{l l *{6}{c}}
    \multicolumn{6}{c}{\bf (b) iUS to \Tone registration } \\
		\toprule
  \rowcolor{white}
		&  & \multicolumn{1}{c}{\bf 0-4 mm } & \multicolumn{1}{c}{\bf 4-8 mm} & \multicolumn{1}{c}{\bf 8-12 mm } & \multicolumn{1}{c}{\bf 12-16 mm} \\ 
		\midrule
\rowcolor{LightGray}
\midrule
\multicolumn{2}{l}{Acquired iUS}  & 12.6 (3.2) & 12.9 (3.7) & 14.6 (3.8) & 14.9 (4.0) \\
\midrule
 &MM-GAN~\cite{Sharma20}& 6.3 (3.7) & 7.7 (5.2) & 7.6 (5.1) & 11.0 (5.2) \\
 \rowcolor{LightGray}
 &ResViT~\cite{resvit}& \textbf{2.2 (2.4)} & \textbf{4.0 (4.3)} & \textbf{4.1 (4.3)} & \textbf{7.9 (5.7)} \\
 & \revmod{M2DN~\cite{meng2024multi}}  & 5.4 (2.9) & 6.6 (3.8) & 9.8 (4.4) & 12.2 (4.0) \\
  \rowcolor{LightGray}
\multirow{-4}{*}{\rotatebox[origin=c]{90}{\textbf{Tran. \Tone}} } &Ours &  \textbf{2.3 (1.7)} & \textbf{2.8 (2.4)} & \textbf{3.2 (3.3)} & \textbf{6.5 (5.0)} \\
\arrayrulecolor{black}\bottomrule
	\end{tabular}
 }
  	\resizebox{0.32\textwidth}{!}{
 \addtolength{\tabcolsep}{-0.5em}
	\begin{tabular}{l l *{6}{c}}
    \multicolumn{6}{c}{\bf (c) iUS to FLAIR registration } \\
		\toprule
  \rowcolor{white}
		&  & \multicolumn{1}{c}{\bf 0-4 mm } & \multicolumn{1}{c}{\bf 4-8 mm} & \multicolumn{1}{c}{\bf 8-12 mm } & \multicolumn{1}{c}{\bf 12-16 mm} \\ 
		\midrule
\rowcolor{LightGray}
\midrule
\multicolumn{2}{l}{Acquired iUS}  & 8.9 (2.6) & 9.2 (3.0) & 9.4 (3.2) & 12.6 (4.4) \\
\midrule
 &MM-GAN~\cite{Sharma20}& 7.5 (4.1) & 8.5 (4.0) & 10.4 (4.1) & 12.2 (4.3) \\
 \rowcolor{LightGray}
 &ResViT~\cite{resvit} & \textbf{1.9 (2.7)} & \textbf{3.4 (4.6)} & \textbf{5.1 (5.0)} & 10.8 (6.2) \\
 & \revmod{M2DN~\cite{meng2024multi}}  & 4.6 (3.5) & 7.1 (4.7) & 9.4 (4.7) & 13.2 (5.2) \\
 \rowcolor{LightGray}
\multirow{-4}{*}{\rotatebox[origin=c]{90}{\textbf{Tran. Flair}} } &Ours & \textbf{1.8 (1.2)} & \textbf{2.3 (2.1)} & \textbf{4.4 (4.7)} & \textbf{8.0 (6.5) }  \\
\arrayrulecolor{black}\bottomrule
	\end{tabular}
 }
\end{table*}

\subsubsection{Fusing operation}
The ablation study results in Table~\ref{tab:Scores_ablation_all} demonstrate the effectiveness of our proposed fusion operation compared to two commonly used techniques for handling missing data: \textbf{Concat. w/ zeros} and \textbf{Average}. 

The \textbf{Concat. w/ zeros} method, which concatenates available images with zero imputation for missing modalities, obtained lower performance, especially in terms of perceptual quality. For instance, it achieves a PSNR of 26.1 dB, an LPIPS of 15.6\% for \iUS, and a PSNR of 24.0 dB with an LPIPS of 8.7\% for FLAIR. 

The \textbf{Average} method, which averages features extracted independently from each available image at each level of the hierarchy, performs slightly better, especially in terms of SSIM and PSNR across modalities. For example, it achieves a PSNR of 27.0 dB and an SSIM of 89.0\% for \Ttwo, indicating that extracting features from each image independently helps retain more information across missing modalities. 

Our proposed fusion operation outperforms both methods across almost all metrics and modalities. For instance, it achieves a PSNR of 27.0 dB for \iUS, 28.6 dB for \Ttwo, 29.8 dB for \Tone, and 27.4 dB for FLAIR. The SSIM scores are also the highest (e.g., 91.7\% for \Tone and 91.0\% for FLAIR), while the LPIPS scores are the lowest. These results show that our probabilistic fusion approach, which integrates information more effectively across available modalities, leads to more accurate and visually realistic image synthesis, particularly compared to more straightforward concatenation or averaging methods.

\subsubsection{GAN regularization}
Another key component of our framework is the use of the GAN loss. We present results with ($\lambda_{\text{GAN}}=0.025$) and without ($\lambda_{\text{GAN}}=0$) the GAN loss in Figure~\ref{fig:qualitative_results_synthesis} and Table~\ref{tab:Scores_ablation_all}. As shown in Figure~\ref{fig:qualitative_results_synthesis}, using the GAN loss improves the visual quality of the generated images. Without the GAN loss, our model tends to produce smoother and less realistic images, particularly lacking the speckles characteristic of ultrasound images. These qualitative improvements align with the perceptual results in Table~\ref{tab:Scores_ablation_all}, where lower LPIPS scores are achieved with the GAN loss. However, we also found that the GAN loss leads to lower SSIM and PSNR scores. This illustrates the typical challenge in assessing image synthesis methods, where metrics may not agree. 

This challenge motivated us to conduct additional experiments to further evaluate the quality of the synthesized images in two downstream tasks, where errors can be quantified with actionable and interpretable metrics: image segmentation and registration.

\begin{table}[t]
	\centering
	
	\caption{\textbf{Comparison against the state-of-the-art unified models on brain tumor segmentation downstream tasks in \iUS.} Best performance in bold is based on a paired Wilcoxon signed-rank test ($p<0.05$).}\label{tab:Scores_segmentation}
 	\resizebox{0.49\textwidth}{!}{
 \addtolength{\tabcolsep}{-0.5em}
	\begin{tabular}{l c *{4}{c}}
		\toprule
  \rowcolor{white}
    & \multicolumn{2}{c}{RESECT-SEG~\cite{resect-seg}} & \multicolumn{2}{c}{ReMIND~\cite{remind}} \\
    \cmidrule(lr){2-3} \cmidrule(lr){4-5} 
		 & Dice Score ($\%$)$\uparrow$ & ASSD (mm)$\downarrow$ & Dice Score ($\%$)$\uparrow$ & ASSD (mm)$\downarrow$ \\
   \midrule
   \rowcolor{LightGray}
MM-GAN~\cite{Sharma20} & 53.3 [37.1 - 69.8] & 4.1 [3.4 - 5.3] &  57.9 [52.6 - 71.5] & 5.0 [3.3 - 6.0] \\
ResViT~\cite{resvit} & \textbf{67.3 [48.3 - 80.5]} & \textbf{2.3 [1.7 - 3.7]} &  \textbf{74.7 [69.0 - 79.6]} & \textbf{2.6 [2.3 - 3.3]} \\
    \rowcolor{LightGray}  \revmod{M2DN~\cite{meng2024multi}}  & 65.2 [49.2 - 76.6] & 3.0 [2.4 - 4.6] & 62.2 [49.2 - 63.3] & 4.6 [4.2 - 5.7] \\
    Ours & \textbf{73.6 [54.4 - 81.3]} & \textbf{2.3 [1.5 - 4.0]} & \textbf{77.6 [67.6 - 84.4]} & \textbf{2.4 [1.7 - 3.6]}
 \\
    \midrule
    \rowcolor{LightGray}
    Fully~\cite{resect-seg} & - & - & \textbf{73.4 [67.5 - 75.7]} & \textbf{2.4 [2.2 - 2.4]} \\
    \midrule
    Expert &  - &  -  & 84.2 [83.3 - 84.8] &1.5 [1.0 - 1.6]\\
\arrayrulecolor{black}\bottomrule
	\end{tabular}
 }
\end{table}

\subsection{Learning brain tumor Segmentation in \iUS from synthetic iUS}

The results of the downstream task of brain tumor segmentation in ultrasound images are presented in Table~\ref{tab:Scores_segmentation}. All nnU-Net models were trained using synthetic ultrasound images generated by each cross-modal synthesis method from the same MR dataset. This MR dataset contains all combinations of possible input MR data (T1, T2, FLAIR, T1+T2, etc). Our proposed framework demonstrated superior performance at synthesizing ultrasound data compared to competing approaches, achieving the highest Dice scores on both the RESECT-SEG (73.6\%) and ReMIND (77.6\%) datasets and significantly outperforming MM-GAN and \revmod{M2DN}. These results show that the ultrasound images synthesized by our method from MR data are accurate, particularly in regions in and near the tumor.
   \revmod{Notably, our approach reached comparable performance to an ensemble of nnU-Net models trained using a small annotated dataset of real iUS, achieving a higher median Dice score ($77.6\%$ vs $73.4\%$), while maintaining the same median ASSD ($2.4$ mm). This highlights not only the quality of the generated images but also their utility for training segmentation models in a new modality, such as ultrasound, without requiring additional annotations.} 
Finally, our approach achieves performance close to that of expert neurosurgeon, while enabling rapid 3D iUS delineation in under 10 seconds compared to experienced neurosurgeons (32-93 minutes).

\begin{figure*}[tb!]
    \centering
    \includegraphics[width=0.78\textwidth]{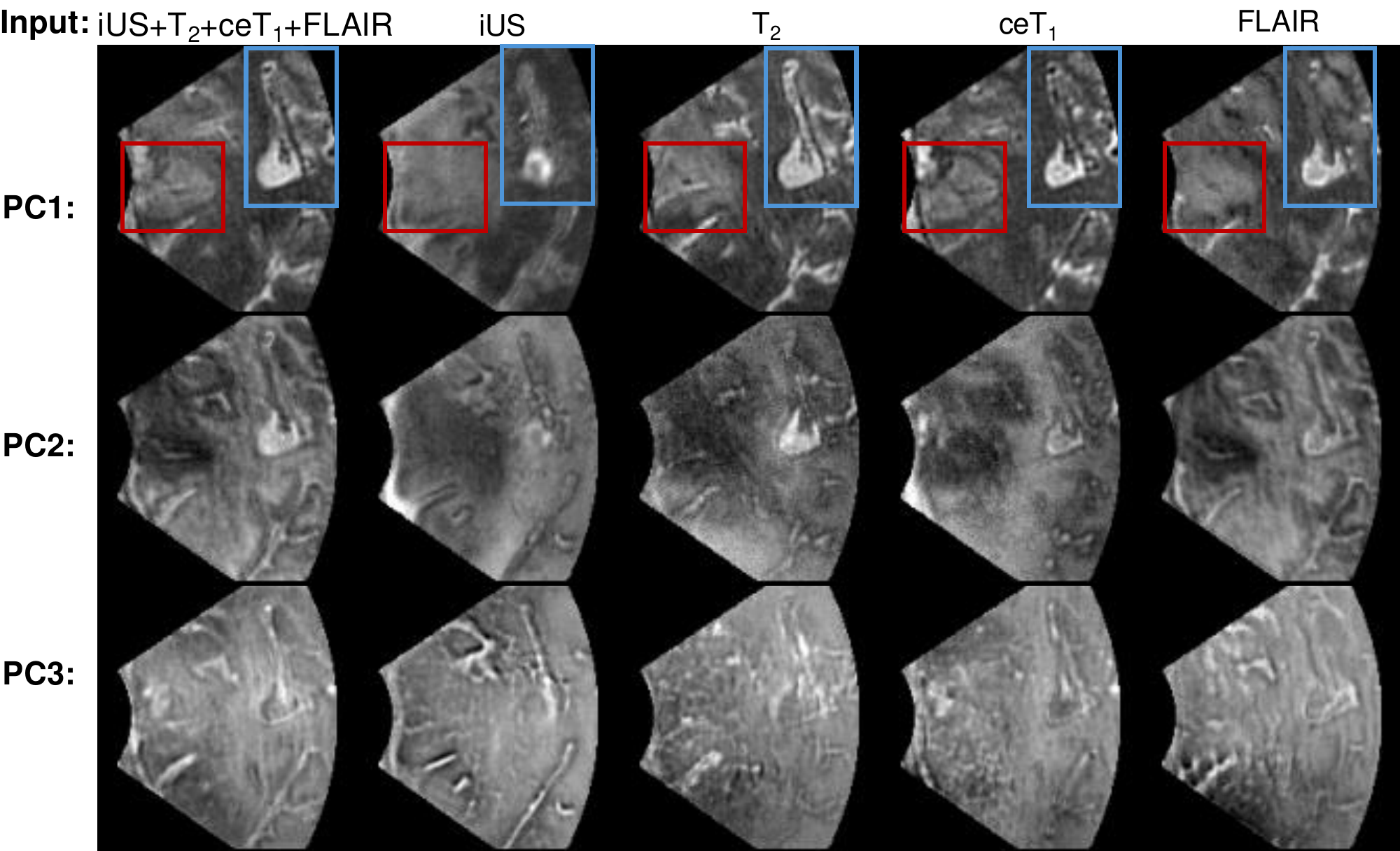}
    \caption{Principal Component Analysis on the first three components (PC1: $39\%$, PC2: $16\%$, PC3: $13\%$) of the latent variable at the highest level $z_1$ estimated using (a) iUS+\Ttwo+\Tone+FLAIR; (b) iUS; (c) \Ttwo (d) \Tone (e) FLAIR as input. Similar representations are obtained for all combinations, in particular in the tumor region (red) and around the ventricle (blue).}
    \label{fig:latent_representation}
\end{figure*}
\begin{figure*}[tb!]
    \centering
    \includegraphics[width=0.78\textwidth]{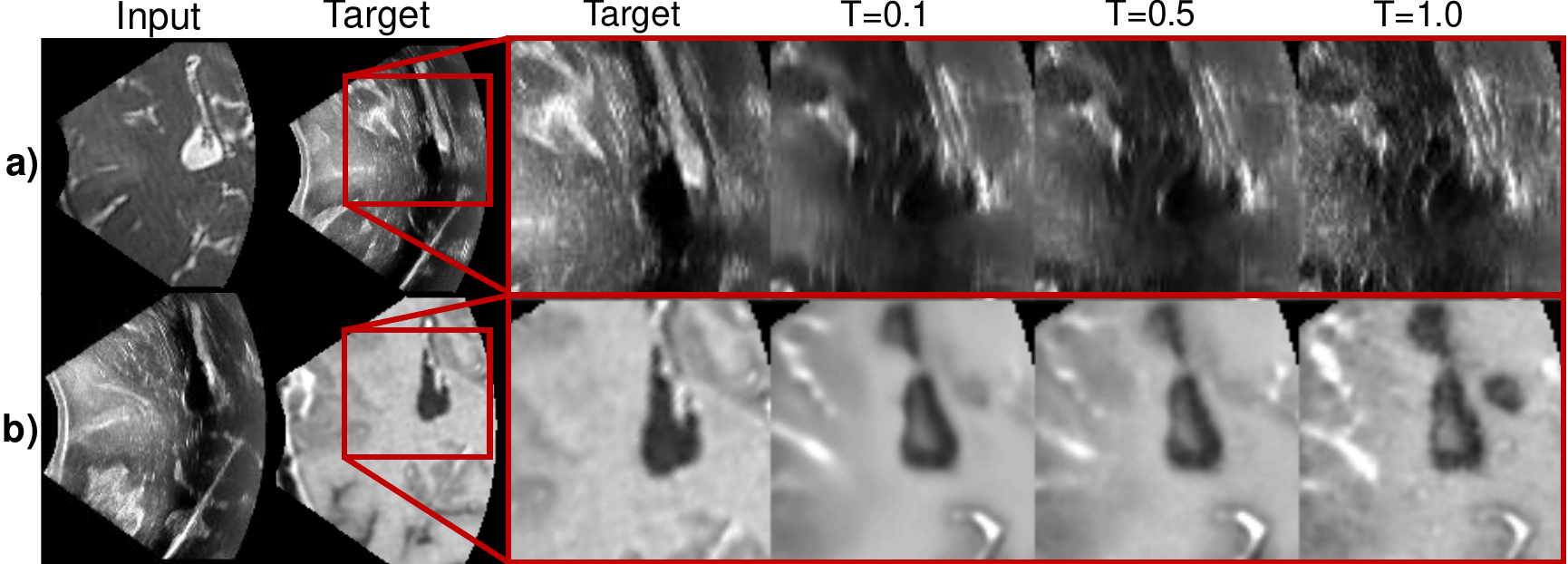}
    \caption{Impact of the temperature $T$ on the quality of the reconstructed images for a) \iUS to \Ttwo; b) \iUS to \Tone synthesis.}
    \label{fig:stochasticity}
\end{figure*}

\subsection{Improving \iUS-MR image registration with synthetic MRI}
The results of the multimodal image registration task between MRI and 3D intraoperative ultrasound (\iUS) images are presented in Table~\ref{tab:Scores_reg_t1}, where various approaches are compared in terms of Target Registration Error (TRE).

We found that performing image registration using synthetic MR data generated from \iUS, rather than acquired 3D \iUS data, significantly improves the registration process. Notably, the complexity of registering \iUS with MRI varies depending on the MR sequence. The registration performs well with \Ttwo images even with large initial displacements (up to $12$ mm). In contrast, for \Tone or FLAIR sequences, the registration often converges to incorrect solutions ($\textgreater 10$ mm), even for small initial displacements ($0$-$4$ mm). Conversely, registering synthetic MR image from our method leads to significant improvements consistently for all MR sequences, even with large initial displacements (up to 12 mm). For instance, the average registration error is reduced from $14.6$ to $3.2$ mm for \Tone in the $8$-$12$ mm displacement range. 
While our approach does not always fully converge to correct solutions for substantial displacements ($12$-$16$ mm), it still reduces the displacement by a factor of two on average. Overall, these findings demonstrate that synthesizing MR images from \iUS significantly improves registration accuracy, particularly for \Tone and FLAIR sequences and larger deformations, where contrast differences often cause conventional registration methods to converge to suboptimal solutions.

We also compared our framework with other synthesis frameworks (MM-GAN, ResViT, and M2DN). Our method consistently outperformed the competing approaches in terms of TRE, particularly for larger displacement ranges. For instance, in the \Ttwo registration, our approach obtained the lowest TRE across all ranges, achieving $2.0$ mm for the $8$-$12$ mm range, compared to $4.4$ mm for MM-GAN, $2.5$ mm for ResViT, and  \revmod{$5.8$ mm for M2DN}. Similarly, for ceT1, our method achieved a TRE of $2.8$ mm for the $4$-$8$ mm range, significantly outperforming MM-GAN ($7.7$ mm),  ResViT ($4.0$ mm), and  \revmod{M2DN ($6.6$ mm)}. Our method's advantage was even more evident for the FLAIR modality, with a TRE of $1.8$ mm for the $0$-$4$ mm range, compared to $7.5$ mm, $1.9$ mm,  \revmod{and $4.6$ mm for MM-GAN, ResViT, and M2DN,} respectively. Overall, these results show our translated MR images better preserve anatomical structures in a similar contrast to the target MR images.

\section{Discussion and Conclusion}
In this section, we discuss some additional components of our framework and conclude this work.
\subsection{Latent Representation Analysis}

This section analyzes the learned latent representation at the pixel level. A key limitation of MVAEs~\cite{wu2018multimodal} is the misalignment of latent representations, where the model tends to learn inconsistent representations for different input data modalities. To address this, we modeled the variational posterior as a mixture of product-of-experts. As shown in Appendix 1.2, this mixture encourages the approximate posterior distribution from incomplete data to align with the true posterior from complete data, thereby estimating the missing information and ensuring consistency across the different input modalities.

To experimentally assess the alignment of the latent representations, we performed a Principal Component Analysis (PCA) on the latent variable $z_1$ from various input combinations (iUS, \Ttwo, \Tone, or FLAIR) and from all modalities combined. The PCA was conducted by randomly selecting $N=10,000$ pixel representations from all slices of a patient and all combinations of input data. The results, presented in Figure~\ref{fig:latent_representation}, show a good alignment between the latent representations across different modalities despite the inherent differences in contrast and information between them. Interestingly, the first principal component of the $z_1$ obtained using all images presents a clear contrast between ventricles and white matter, as in a \Ttwo image, and clear delineation of the tumor, as in a \Tone image. While this information is not present in the \iUS, its latent representation obtained from \iUS tries to estimate it. These results highlight the effectiveness of our mixture of product-of-experts approach, 
as it successfully produces well-aligned latent representations for different inputs.

\subsection{Variability in the samples} It is common to lower the temperature of the conditional distributions when sampling from HVAEs on challenging datasets~\cite{vahdat2020nvae}. This is done by scaling down the standard deviation of the Normal distributions at each level in the approximate posterior. It often improves the samples' quality but also reduces their diversity. Figure~\ref{fig:stochasticity} shows how the cross-modal synthesis quality and diversity vary with temperature. We found that sampling with a low temperature tends to produce smooth images that accurately reflect the overall structure of the images. However, these images are less realistic due to their lack of finer, modality-specific details. In contrast, higher temperatures introduce modality-specific features, such as speckles in synthetic ultrasound from \Ttwo or contrast agents in contrast-enhanced T1 images from \iUS, which correspond to information not explicitly present in the conditioning image. This indicates that modality-specific details are captured in the variability of the Normal distribution. By adjusting the temperature, we can modulate how much of this modality-specific information is incorporated into the synthesized images, with higher temperatures allowing for the generation of more realistic but less reliable images. We found that a temperature of $0.5$ was a good compromise for the realism and variability of samples.

\subsection{Limitations} While MMHVAE demonstrates promising results for medical multimodal synthesis, several limitations should be acknowledged. First, the framework has been primarily evaluated on imaging datasets, and its applicability to non-imaging modalities (e.g., genomics or clinical data) remains to be established. Second, the method relies on the availability of sufficiently paired multimodal data during training; in extreme cases where a modality is consistently missing or available in isolation, performance is expected to degrade. Finally, the interpretability of the learned latent representations remains limited, and future work is needed to link the latent factors to clinically meaningful variables.

\subsection{Conclusion}

We proposed a framework for synthesizing missing imaging data across modality, with the flexibility to handle all combination of data missingness at training and inference time. The proposed model relies on a mixture of product-of-experts to encore observed information and estimate missing ones. Our experiment results verified the effectiveness of our model empirically.

\section*{Acknowledgments}
This work was supported by the National Institutes of Health (R01EB032387, R01EB027134, P41EB015902, P41EB028741 and K25EB035166). R.D. received a Marie Skłodowska-Curie grant No 101154248 (project: SafeREG). This work was performed using HPC resources from GENCI–IDRIS (Grant 2024-SafeREG, 2025-SafeREG). The research leading to these results has received funding from the program "Investissements d’avenir" ANR-10-IAIHU-06.

\ifCLASSOPTIONcaptionsoff
  \newpage
\fi

\bibliographystyle{IEEEtran}
\bibliography{references}

\begin{IEEEbiography}[{\includegraphics[width=1in,height=1.25in,clip,keepaspectratio]{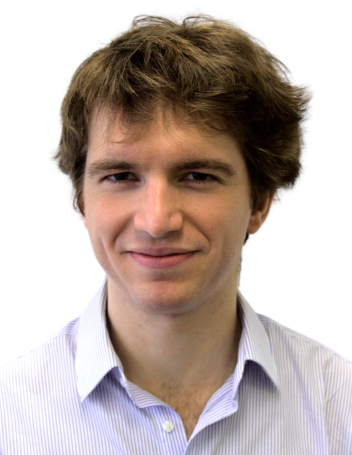}}]{Reuben Dorent} received a PhD degree from King's College London, in 2022. He then joined the  Surgical Planning Laboratory, Brigham and Women’s Hospital, Harvard Medical School. He is now a Marie Skłodowska-Curie fellow at Inria Paris-Saclay and the Paris Brain Institute. His research focuses on medical image segmentation, synthesis, cross-modal domain adaptation, and optimization strategies for missing data in medical contexts.\end{IEEEbiography}

\begin{IEEEbiography}[{\includegraphics[width=1in,height=1.25in,clip,keepaspectratio]{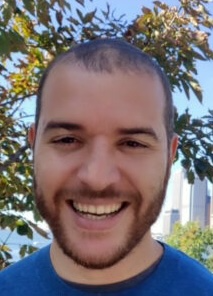}}]{Nazim Haouchine} received his PhD in Computer Science from the University of Lille and INRIA in 2015, in France. He is now an Assistant Professor at Harvard Medical School and a Research Associate at the Brigham and Women's Hospital. His research interests are in building translational technologies for computer-assisted medical interventions, combining research on computer vision and graphics, deep learning, and physics-based simulation. 
\end{IEEEbiography}

\begin{IEEEbiography}[{\includegraphics[width=1in,height=1.25in,clip,keepaspectratio]{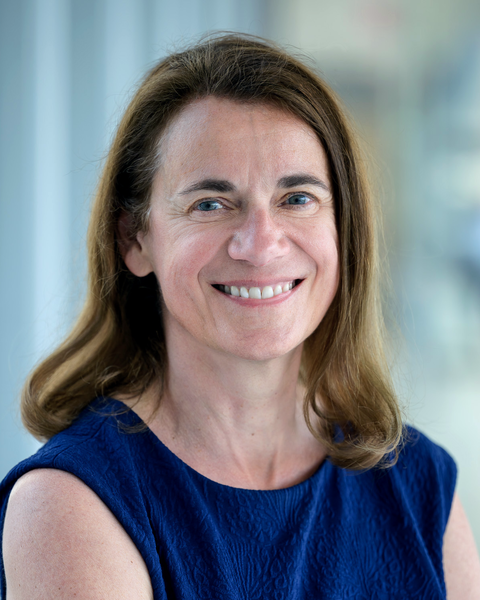}}]{Alexandra J. Golby} received her M.D. from Stanford Medical School. She completed her clinical training in neurosurgery at Stanford and at Brigham and Women’s/Boston Children’s Hospitals. She is a Neurosurgeon at Brigham and Women’s Hospital and a Professor of Neurosurgery and Radiology at Harvard Medical School. Her research focuses on advanced image-guided neurosurgery, functional brain mapping, intra-operative imaging, and minimally invasive approaches.
\end{IEEEbiography}

\begin{IEEEbiography}[{\includegraphics[width=1in,height=1.25in,clip,keepaspectratio]{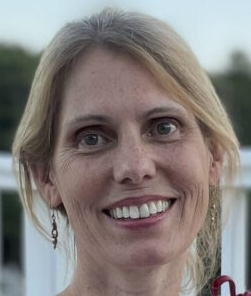}}]{Sarah Frisken} received a PhD degree from Carnegie Mellon University. She was a Distinguished Research Scientist at Mitsubishi Electric Research Labs, Professor of Computer Science at Tufts University, and  CEO of 61 Solutions Inc. She is currently an Associate Professor at Harvard Medical School and Lead Investigator at Brigham and Women’s Hospital. Her research interests include shape representation and modeling, medical image processing, and image-guided neurosurgery.
\end{IEEEbiography}

\begin{IEEEbiography}[{\includegraphics[width=1in,height=1.25in,clip,keepaspectratio]{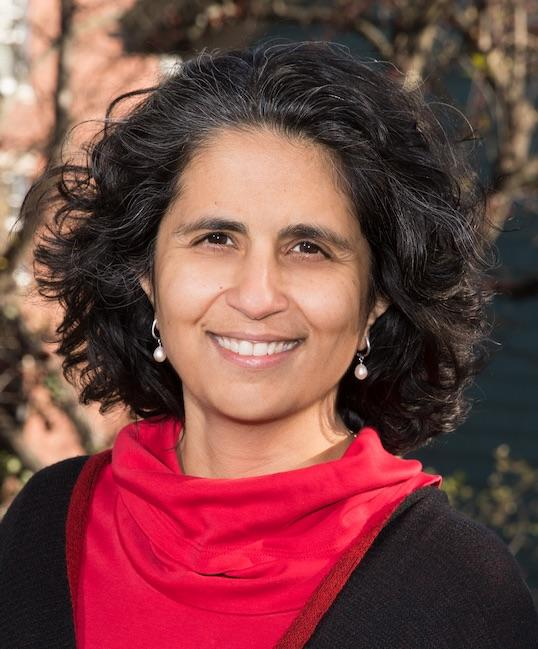}}]{Tina Kapur}
 (Member IEEE) received the SM and PhD degrees in Electrical Engineering and Computer Science from the Massachusetts Institute of Technology. She is Associate Professor of Radiology at Harvard Medical School and Brigham and Women’s Hospital, where she is Lead Investigator and Executive Director of the Image-Guided Therapy Program. Her research interests include image segmentation, registration, surgical navigation, image-guided therapy, and open-source platforms for medical imaging.
\end{IEEEbiography}

\begin{IEEEbiography}[{\includegraphics[width=1in,height=1.25in,clip,keepaspectratio]{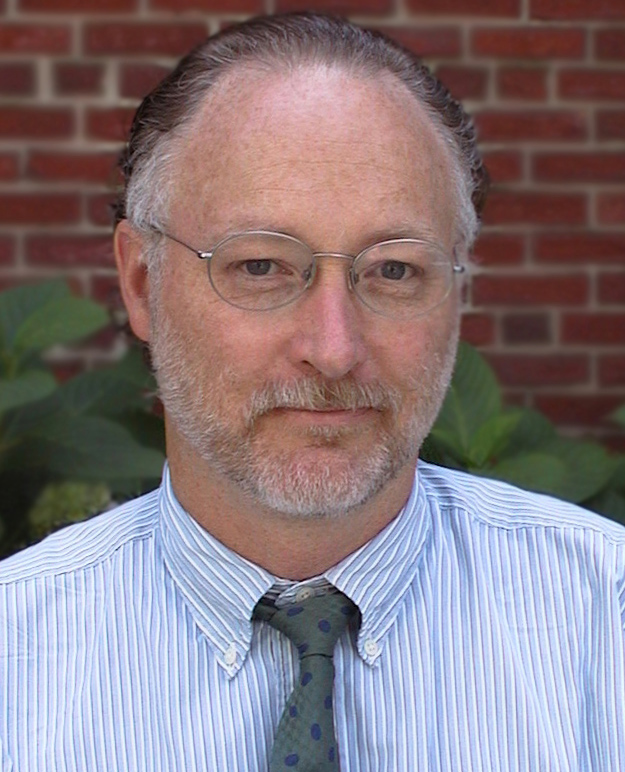}}]{William Wells}  is Professor of Radiology at Harvard Medical School and Brigham and Women's Hospital, a research scientist at the MIT Computer Science and Artificial Intelligence Laboratory, and a
member of the affiliated faculty of the Harvard-MIT division of Health Sciences and Technology.  He received a PhD in computer vision from MIT, and has pursued research in medical image computing at the Surgical Planning Laboratory.  He is widely known for his ground-breaking work on multi-modality registration by maximization of Mutual Information.
\end{IEEEbiography}

\begin{figure*}[th!]
    \centering
    \subfigure[\Ttwo]{\includegraphics[width=0.33\textwidth]{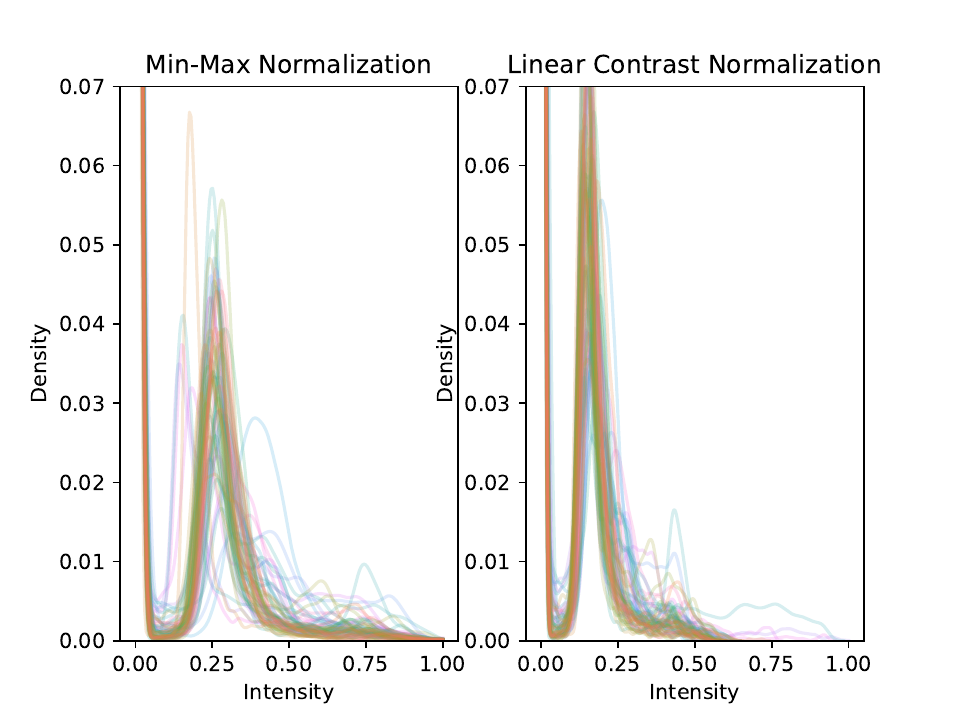}} 
    \subfigure[\Tone]{\includegraphics[width=0.33\textwidth]{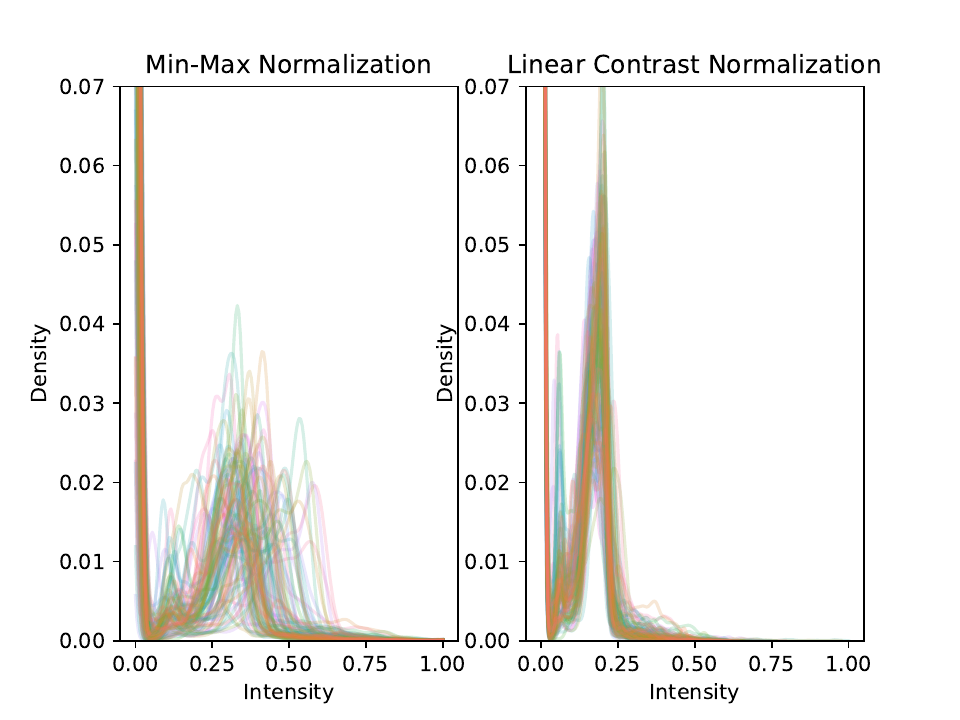}} 
    \subfigure[FLAIR]{\includegraphics[width=0.33\textwidth]{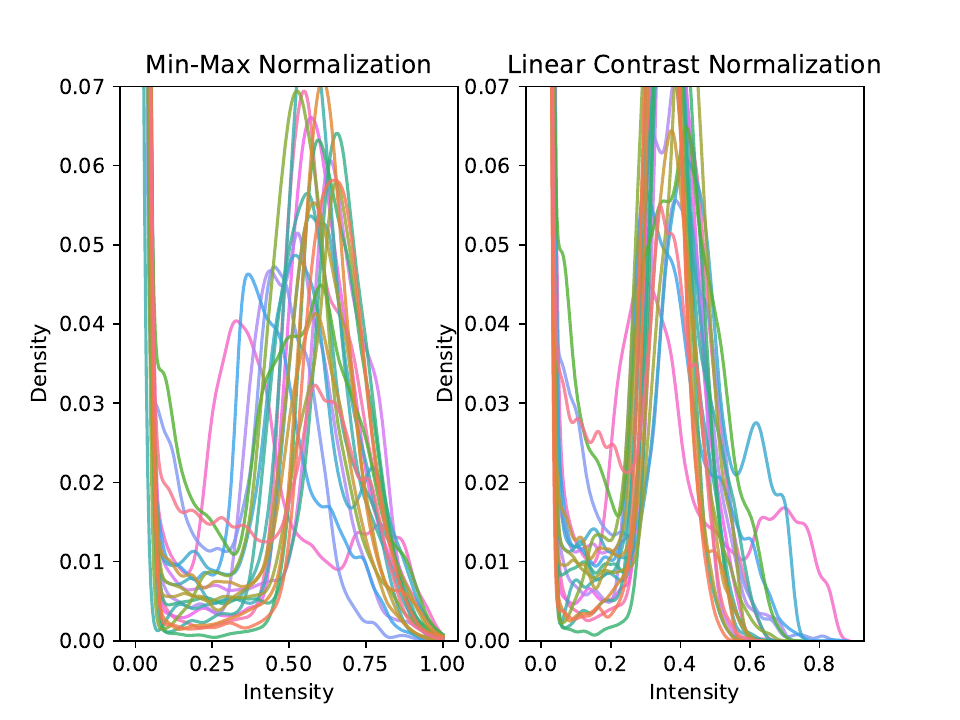}}
    \caption{Intensity distribution of (a) \Ttwo; (b) \Tone; (c) FLAIR images using either min-max normalization and the proposed harmonization technique. }
    \label{fig:distribution}
\end{figure*}

\section{Appendix}
\subsection{Proof ELBO $\mathcal{L}^{\text{ELBO}}_{\text{MMHVAE}}$}\label{appendix:elbo}
Let $\bm{x}^{o}_{\bm{r}}$ be an observed set of images. For any distribution $q_{\phi}(\bm{z}|\bm{x}^{o}_{\bm{r}})$, the evidence $\log p_{\theta}\left(\bm{x}^{o}_{\bm{r}}\right)$ is lower-bounded by the tractable variational ELBO  $\mathcal{L}^{\text{ELBO}}(\bm{x}; \theta, \phi)$:
\begin{multline}
\label{eq:appendix_elbo_general}
\mathcal{L}^{\text{ELBO}}(\bm{x}^{o}_{\bm{r}}; \theta, \phi)=  \sum_{\substack{j=1 \\ \text{s.t. } r_j=1}}^{M}\mathbb{E}_{q_{\phi}(\bm{z}|\bm{x}^{o}_{\bm{r}})}[\log p_{\theta}(x_j|\bm{z})] \\
- \KL\left[q_{\phi}(\bm{z}|\bm{x}^{o}_{\bm{r}})\Vert p_{\theta}(\bm{z})\right] \ .
\end{multline}

In this work, we propose to express the variational $q_{\phi}(\bm{z}|\bm{x}^{o}_{\bm{r}})$  as a mixture (convex combination) of distributions represented by:
\begin{equation}
    q^{\text{MMHVAE}}_{\phi}(\bm{z}|\bm{x}^{o}_{\bm{r}}) = \sum_{\bm{r'}\in S_{\bm{{r}}}} \alpha^{(\bm{r})}_{\bm{r'}} q_{\phi}(\bm{z}|\bm{x}^{o}_{\bm{r'}}) \ .
\end{equation}
Therefore:
\begin{equation}
\label{eq:appendix_elbo}
\mathbb{E}_{q_{\phi}(\bm{z}|\bm{x}^{o}_{\bm{r}})}[\log p_{\theta}(x_j|\bm{z})] = \sum_{\bm{r'}\in S_{\bm{{r}}}} \alpha^{(\bm{r})}_{\bm{r'}} 
\mathbb{E}_{q_{\phi}(\bm{z}|\bm{x}^{o}_{\bm{r'}})}[\log p_{\theta}(x_j|\bm{z})] .
\end{equation}
Moreover, as the KL divergence is convex in the pair of probability measures and $\sum_{\bm{r'}\in S_{\bm{{r}}}} \alpha^{(\bm{r})}_{\bm{r'}}=1$:
\begin{equation}
\label{eq:appendix_convex}
    \KL\left[q_{\phi}(\bm{z}|\bm{x}^{o}_{\bm{r'}})\Vert p_{\theta}(\bm{z})\right] \leq \sum_{\bm{r'}\in S_{\bm{{r}}}} \alpha^{(\bm{r})}_{\bm{r'}} \KL\left[q_{\phi}(\bm{z}|\bm{x}^{o}_{\bm{r'}})\Vert p_{\theta}(\bm{z})\right] 
\end{equation}
Hence, by combining Eq.~\eqref{eq:appendix_elbo_general}, \eqref{eq:appendix_elbo} and \eqref{eq:appendix_convex}, the evidence $\log p(\bm{x}^{o}_{\bm{r}})$ is lower-bounded by:
\begin{equation}
\mathcal{L}^{\text{ELBO}}_{\text{MMHVAE}}(\bm{x}^{o}_{\bm{r}}; \theta, \phi) \triangleq  \sum_{\bm{r'}\in S_{\bm{{r}}}} \alpha^{(\bm{r})}_{\bm{r'}} \mathcal{L}(\bm{x}^{o}_{\bm{r}}, \bm{r'}; \theta, \phi)
\end{equation}
where:
\begin{equation}
\begin{split}
\mathcal{L}(\bm{x}^{o}_{\bm{r}}, \bm{r'}; \theta, \phi) &=  \sum_{\substack{j=1 \\ \text{s.t. } r'_j=1}}^{M}\mathbb{E}_{q_{\phi}(\bm{z}|\bm{x}^{o}_{\bm{r'}})}[\log p_{\theta}(x_j|\bm{z})] \\
& \quad + \sum_{\substack{j=1 \\ \text{s.t. } (1-r'_j)r_j=1}}^{M}\mathbb{E}_{q_{\phi}(\bm{z}|\bm{x}^{o}_{\bm{r'}})}\left[\log p_{\theta}(x_j|\bm{z})\right] \\
& \quad - \KL\left[q_{\phi}(\bm{z}|\bm{x}^{o}_{\bm{r'}})\Vert p_{\theta}(\bm{z})\right] 
\end{split}
\end{equation}

\subsection{Proof maximal value of variational with respect to $\phi$}\label{appendix:maximalvalue}
Let $\bm{x}^{o}_{\bm{r}}$ be an observed set of images. 

\begin{equation}
\begin{split}
\KL & \left[q_{\phi}(\bm{z}|\bm{x}^{o}_{\bm{r'}}) \Vert p_\theta(\bm{z}\vert \bm{x}^{o}_{\bm{r}})\right] \\
&= 
-\int_z q_{\phi}(\bm{z}|\bm{x}^{o}_{\bm{r'}}) \log \frac{p_\theta(z \vert \bm{x}^{o}_{\bm{r}})}{q_{\phi}(\bm{z}|\bm{x}^{o}_{\bm{r'}})} d\bm{z}\\
&= -\int_z q_{\phi}(\bm{z}|\bm{x}^{o}_{\bm{r'}}) \log \frac{p_\theta(\bm{z},\bm{x}^{o}_{\bm{r}})}{q_{\phi}(\bm{z}|\bm{x}^{o}_{\bm{r'}})p_\theta(\bm{x}^{o}_{\bm{r}})} d\bm{z} \\
&= - \int_z q_{\phi}(\bm{z}|\bm{x}^{o}_{\bm{r'}}) \log \frac{p_\theta\left(\bm{x}^{o}_{\bm{r}} \vert \bm{z}\right)p_{{\theta}}(\bm{z})}{q_{\phi}(\bm{z}|\bm{x}^{o}_{\bm{r'}})} d\bm{z} + \log p_\theta(\bm{x}^{o}_{\bm{r}}) \\
&= -\mathbb{E}_{\bm{z} \sim q_\phi \left(z \vert \bm{x}^{o}_{\bm{r'}}\right)} \left[p_\theta\left(\bm{x}^{o}_{\bm{r}} \vert \bm{z}\right)\right] + \KL \left[q_\phi \left(\bm{z} \vert \bm{x}^{o}_{\bm{r}}\right) \Vert p_{\theta}(\bm{z})\right] \\
& \quad \quad + \log p_\theta(\bm{x}^{o}_{\bm{r}})\\
&=  \log p_\theta(\bm{x}^{o}_{\bm{r}})-\mathcal{L}(\bm{x}^{o}_{\bm{r}}, \bm{r'}; \theta, \phi) 
\end{split}
\end{equation}

Consequently, as $\sum_{\bm{r'}\in S_{\bm{{r}}}} \alpha^{(\bm{r})}_{\bm{r'}}=1$, we can find the gap between the evidence $\log p_\theta(\bm{x}^{o}_{\bm{r}})$ and the ELBO $\mathcal{L}^{\text{ELBO}}(\bm{x}^{o}_{\bm{r}}; \theta, \phi)$: 

\begin{equation}
\begin{split}
    \log p_\theta(\bm{x}^{o}_{\bm{r}})- & \mathcal{L}^{\text{ELBO}}(\bm{x}^{o}_{\bm{r}}; \theta, \phi) \\
    &= \sum_{\bm{r'}\in S_{\bm{{r}}}} \alpha^{(\bm{r})}_{\bm{r'}} \KL\left[q_{\phi}(\bm{z}|\bm{x}^{o}_{\bm{r'}}) \Vert p_\theta(\bm{z} \vert \bm{x}^{o}_{\bm{r}})\right]
\end{split}
\end{equation}

Consequently, as $\alpha^{(\bm{r})}_{\bm{r'}}\geq 0$ the maximal value of the ELBO $\mathcal{L}^{\text{ELBO}}(\bm{x}^{o}_{\bm{r}}; \theta, \phi)$ with respect to $\phi$ is reached when:

\begin{equation}
    \forall \bm{r'}\in S_{\bm{{r}}}, \ q_{\phi}(\bm{z}|\bm{x}^{o}_{\bm{r'}}) = p_\theta(\bm{z} \vert \bm{x}^{o}_{\bm{r}})
\end{equation}

\subsection{Proof posterior parameterization}\label{appendix:parametrization}
Using Bayes' rule, the true conditional distribution $p_{\theta}(\bm{z}\vert\bm{x}^o_{\bm{r}})$ can be written as: $p_{\theta}(\bm{z}\vert\bm{x}^o_{\bm{r}})=\frac{p_{\theta}(\bm{z})}{p(\bm{x}^o_{\bm{r}})}p(\bm{x}^o_{\bm{r}}|\bm{z})$, where $p(\bm{x}^o_{\bm{r}}|\bm{z}) = \prod_{\substack{j=1  \text{ s.t. } r_j=1}}^M p(x_j|\bm{z})$ (Eq.~\ref{eq:graphicalmodel}). Consequently:
\begin{equation}
    \label{eq:proof1_eq1}
    p_{\theta}(\bm{z}|\bm{x}^o_{\bm{r}}) \propto \prod_{\substack{j=1 \\ \text{s.t. } r_j=1}}p(x_{j}|\bm{z})
\end{equation}
Then, the marginal likelihoods $p_{\theta}(x_j|\bm{z})$ can be factorized as:

\begin{equation}
    \label{eq:proof1_eq2}
    \begin{split}
         p_{\theta}(x_j|\bm{z})&=\frac{p_{\theta}(x_j,z_1,..,z_L)}{p_{\theta}(z_1,..,z_L)}\\
                         &= \frac{p_{\theta}(x_j,\bm{z_{>1}})}{p_{\theta}(\bm{z_{>1}})} \frac{p_{\theta}(z_1|x_j,\bm{z_{>1}})}{p_{\theta}(z_1|\bm{z_{>1}})} \\
                         &= \dotsc\\
                         &= \frac{p_{\theta}(z_L,x_j)}{p(z_L)} \prod_{l=1}^{L-1} \frac{p_{\theta}(z_l|x_j,\bm{z_{>l}})}{p_{\theta}(z_l|\bm{z_{>l}})} \\
                         &= p_{\theta}(x_j)\frac{p_{\theta}(z_L|x_j)}{p(z_L)} \prod_{l=1}^{L-1} \frac{p_{\theta}(z_l|x_j,\bm{z_{>l}})}{p_{\theta}(z_l|\bm{z_{>l}})} .
    \end{split}
\end{equation}
Consequently, combining Eq.~\eqref{eq:proof1_eq1} and Eq.~\eqref{eq:proof1_eq2}, the true conditional distribution $p_{\theta}(\bm{z}|\bm{x}^o_{\bm{r}})$ can be factorized as:
\begin{multline}
    p_{\theta}(\bm{z}|\bm{x}^o_{\bm{r}})\propto p(z_L)\prod_{l=1}^{L-1}p_{\theta}(z_l|\bm{z_{>l}})\\
    \quad  \prod_{\substack{j=1 \\ \text{s.t. } r_j=1}}\left[\frac{p_{\theta}(z_L|x_j)}{p(z_L)} \prod_{l=1}^{L-1} \frac{p_{\theta}(z_l|x_j,\bm{z_{>l}})}{p_{\theta}(z_l|\bm{z_{>l}})}\right],
\end{multline}
which can be rewritten as:
\begin{multline}
    p_{\theta}(\bm{z}|\bm{x}^o_{\bm{r}})=  \left( p(z_L)\prod_{\substack{j=1 \\ \text{s.t. } r_j=1}}^{M}\frac{p_{\theta}(z_L|x_j)}{p(z_L)}\right) \\
    \prod_{l=1}^{L-1}\left(p_{\theta}(z_l|\bm{z_{>l}})\prod_{\substack{j=1 \\ \text{s.t. } r_j=1}}^{M}  \frac{p_{\theta}(z_l|x_j,\bm{z_{>l}})}{p_{\theta}(z_l|\bm{z_{>l}})}\right) .
\end{multline}

\end{document}